\documentclass[runningheads]{llncs}

\usepackage{eccv}

\usepackage{eccvabbrv}
\usepackage{graphicx}
\usepackage{booktabs}
\usepackage{xcolor,colortbl}

\usepackage[accsupp]{axessibility}  

\usepackage{hyperref}

\begin{document}
\interfootnotelinepenalty=10000
\setlength{\textfloatsep}{0.5cm}
\title{RING-NeRF : Rethinking Inductive Biases for Versatile and Efficient Neural Fields } 


%
%
%

%




\titlerunning{RING-NeRF}

\author{Doriand Petit\inst{1,2} \and
Steve Bourgeois\inst{1} \and
Dumitru Pavel\inst{1} \and
Vincent Gay-Bellile\inst{1} \and
Florian Chabot\inst{1} \and
Loïc Barthe\inst{2}}

\authorrunning{D. Petit et al.}

\institute{Université Paris-Saclay, CEA, List,
F-91120, Palaiseau, France \and
IRIT, Université Toulouse III, CNRS, France
}

\maketitle

\begin{center}
    \captionsetup{type=figure}
    \includegraphics[width=0.93\textwidth]{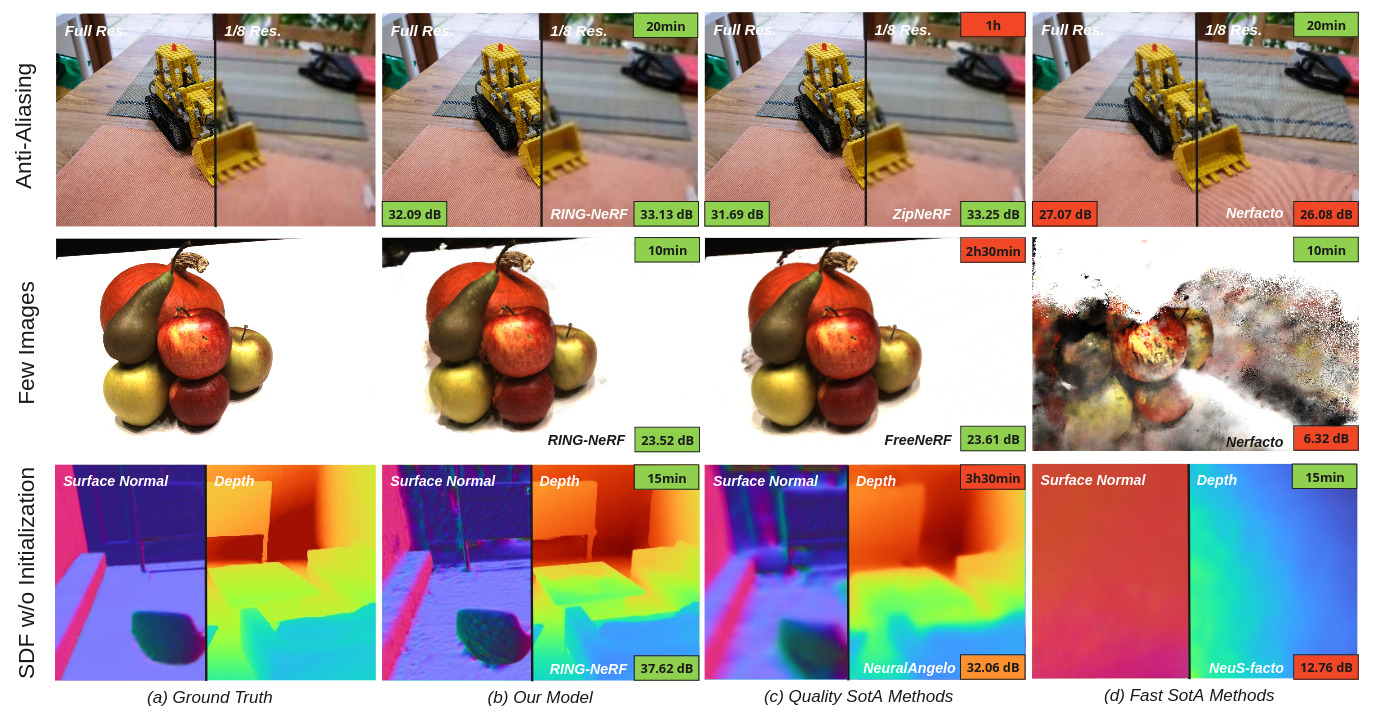}
    \captionof{figure}{RING-NeRF is a simple and versatile architecture which tackles many NeRF common issues  such as robustness to distance of observation, few view supervision and lack of scene-specific initialization for SDF-based reconstruction. It provides on-par performances in terms of quality with SotA dedicated solutions \cite{barron2023zip,yang2023freenerf,li2023neuralangelo} and  and in terms of efficiency with fast methods \cite{tancik2023nerfstudio,Yu2022SDFStudio}. 
}

    \label{fig:results}
\end{center}



\begin{abstract}
Recent advances in Neural Fields mostly rely on developing task-specific supervision which often complicates the models. Rather than developing hard-to-combine and specific modules, another approach generally overlooked is to directly inject generic priors on the scene representation (also called inductive biases) into the NeRF architecture. Based on this idea, we propose the RING-NeRF architecture which includes two inductive biases : a continuous multi-scale representation of the scene and an invariance of the decoder's latent space over spatial and scale domains. We also design a single reconstruction process that takes advantage of those inductive biases and experimentally demonstrates on-par performances in terms of quality with dedicated architecture on multiple tasks (anti-aliasing, few view reconstruction, SDF reconstruction without scene-specific initialization) while being more efficient. Moreover, RING-NeRF has the distinctive ability to dynamically increase the resolution of the model, opening the way to adaptive reconstruction. Project page can be found at : \url{https://cea-list.github.io/RING-NeRF} 

\end{abstract}


\section{Introduction}
\label{sec:intro}

Neural Radiance Fields (NeRF) have emerged as a novel method for representing 3D scenes using neural networks. In its original design \cite{mildenhall2021nerf}, a simple multi-layer perceptron (MLP) is trained to reproduce a continuous 5D function that outputs the density and radiance emitted in each direction ($\theta$, $\phi$) at each point (x, y, z) in space. This approach inspired many works due to the impressive quality of its novel view synthesis as well as its simplicity.
However, the so-called "vanilla" NeRF architectures converge very slowly as they use large deep neural networks. Instant-NGP \cite{muller2022instant} introduced a new architecture based on a hierarchical hash-grid pyramid of learnable feature codes describing the 3D scene and combined with a much shallower MLP, called decoder, that transforms the concatenation of the codes, interpolated from the grids, into density and radiance. Resulting in local updates, this method reduced the training duration from hours to minutes.

The great majority of current solutions are now based on these two standard architectures, and many of them are focused on overtaking their associated limitations in terms of: 
\\\textbf{ Nature of scene} - by transitioning from object-centric scenes to open unbounded scenes, using mostly space contraction \cite{barron2022mip};
\\ \textbf{Robustness} - by managing free motion trajectories with variations of the observation distance while avoiding aliasing artefacts, mostly through the integration of Level of Detail (LOD) in the model \cite{barron2021mip,barron2023zip}; or by reducing drastically the number of supervised views through different kinds of regularization \cite{yang2023freenerf,jain2021putting,niemeyer2022regnerf};
\\ \textbf{Extensibility} - by shifting from a holistic and fixed reconstruction process to an incremental (extensibility in the number of views) and adaptive (extensibility in resolution) process, mainly through the use of a frozen decoder \cite{zhu2022nice,liu2020neural}. 

However, most solutions solely focus on one of these limitations and introduce specific and complex mechanisms that both increase the computational cost and lessen the possibilities of combination. The ability to solve jointly these main issues is however essential in real-world applications which often require a robust and extensible reconstruction in an unknown and unbounded environment. 

In this article, rather than introducing yet another heavy and task-specific solution, we propose RING-NeRF, a versatile and simple NeRF architecture by rethinking usual grid-based models to introduce two inductive biases. We first represent a 3D scene as a continuous multi-scale representation and also make the decoder's latent space invariant in position and scale. Together, these two priors enable the production of intrinsic continuous LOD of the scene without explicit supervision. We demonstrate experimentally that, when combined with adapted cone casting and coarse-to-fine optimization, the resulting architecture is able to compete on several tasks with on-par quality performances with dedicated state-of-the-art solutions while improving speed, robustness and extensibility. The overall process is also simple, easy to implement and generic enough to be coupled with specific solutions. Our contributions can be summarized as follows: 
%

\begin{enumerate}
    \item an architecture that, by construction, represents the scene with a continuous and unbounded level of detail without the need for LOD-specific supervision and which permits resolution-adaptive reconstruction;
    \item a distance-aware forward mapping compatible with scene contraction, that takes benefits of the continuous multi-scale representation with an adapted cone casting process to avoid aliasing artefacts when varying the observation distance.
    \item a continuous coarse-to-fine reconstruction process that improves the convergence and stability (especially in challenging setups such as supervision with few viewpoints or no scene-specific initialization for SDF reconstruction).
\end{enumerate}

\section{Related Work}
\label{sec:sota}

From its original iteration \cite{mildenhall2021nerf}, a majority of current research focuses on overtaking limitations of NeRF-based reconstruction in terms of adaptability to various natures of scene, robustness (to varying observation distances or limited amount of viewpoints) and reconstruction extensibility. 
\\ \textbf{Adaptability to various natures of the scene.} The ability of Neural Fields to reconstruct various natures of scene depends on three factors. 
The first one is related to its architecture itself. Tri-plane architectures \cite{hu2023tri,zhuang2023anti,chen2022TensoRF} are mostly designed for object-centric reconstruction (as they provide a higher density of information in the center of the scene) while vanilla and 3D grid-based NeRF are able to cover a wider variety of scenes, though they initially were still constrained to a limited volume. 
A second factor is related to the representation of the 3D space of the scene, especially to represent distant elements in open scenes. Some approaches use two different NeRF models to reconstruct separately the foreground and the background \cite{zhang2020nerf++}, whereas others apply space contraction to the 3D scene coordinates \cite{zhang2020nerf++,barron2022mip,barron2023zip} to map the infinite scene volume into a bounded one.
The last factor is related to the initialization of the Neural Fields. While the random initialization of density-based NeRF can adjust to almost any nature of scenes, the convergence of SDF-based (Signed Distance Function) Neural Fields is extremely sensitive to their parameters' initialization, as stated in \cite{atzmon2020sal}. Current solutions rely on a scene-specific initialization (using an SDF field representing a sphere for outdoor scenes or an inverted sphere for indoor scenes), making them unable to adapt automatically to any scene.
\\ \textbf{Robustness to observation distance variations.}
 The initial NeRF model, as well as most of the subsequent works, relies implicitly on the hypothesis of a constant distance of observation to the scene. 
 Indeed, the NeRF model provides a per-3D-point scene density representation and a rendering process which does not take into account the increasing volume covered by a pixel with respect to the distance to the cameras. As underlined in Mip-NeRF \cite{barron2021mip}, this discrepancy induces artefacts such as over-contrasted images or aliasing phenomenon when the distance of observation differs from the ones used at reconstruction time.
To avoid these artefacts, the rendering process needs to assess the density and color for a volume instead of a point. In the current state-of-the-art, two main approaches are used (or their combination, as in Zip-NeRF \cite{barron2023zip}). The first solution consists of representing the scene with different levels of detail (LOD), in order to vary the precision of the reconstruction based on the observation distance. This is usually done by using a LOD-aware latent space \cite{dou2023multiplicative,takikawa2021neural,turki2023pynerf,barron2021mip,barron2023zip,xu2023VR-NeRF}, meaning that the LOD information is already encoded in the inputs of the MLP. This can be achieved by using a per-LOD decoder as in PyNeRF \cite{turki2023pynerf} or by incorporating LOD information in the latent feature as done in Zip-NeRF \cite{barron2023zip} and VR-NeRF \cite{xu2023VR-NeRF}. One main flaw of these solutions is that they require the supervision of every used LOD which makes it impossible to vary the observation distances between the train views and novel synthesised views. 
A second approach consists of defining the latent representation of a volume as the mean of the latent features of the points included in the volume. It requires integrating the features over the volume, which can be achieved through convolutions for tri-plan representation \cite{hu2023tri,zhuang2023anti}, or through super-sampling of the latent space for 3D grids as also done in Zip-NeRF \cite{barron2023zip}. However, these approaches lengthen the training and rendering processes as they increase the number of computations.
\\ \textbf{Robustness to limited amount of viewpoints.}
By construction, NeRF is subject to the shape-radiance ambiguity \cite{zhang2020nerf++}. If not enough supervision viewpoints are available, the optimization might overfit them while not providing consistent 3D reconstruction nor generalization to non-supervised viewpoints.  
A first family of solutions to overcome this limitation consists of regularizing the reconstruction process through additional losses, whether via geometric  \cite{niemeyer2022regnerf,yu2022monosdf,wynn2023diffusionerf} or semantic \cite{jain2021putting} regularization.
The second family of solutions relies on a progressive reconstruction of the details of the scene, as introduced in FreeNeRF \cite{yang2023freenerf} and Nerfies \cite{park2021nerfies}. Restraining the ability of the model to reconstruct a complex scene at early stages enforces the consistency of the reconstruction over the different supervision viewpoints. This presents the advantage of bringing stability while keeping a fairly simple training process. However, these latter solutions mostly rely on Vanilla models for stability purposes and require long training duration.
%
%
\\ \textbf{Reconstruction extensibility.} Two kinds of extensibility should be distinguished: extensibility with respect to the number of views or the resolution. The first one is related to reconstructing new scene areas while keeping the previously reconstructed ones unchanged. The usual solution consists of using grid-based approaches with a pre-trained position-invariant decoder that is frozen during the reconstruction\cite{zhu2022nice}.
 On the other side, the extensibility of the resolution consists in dynamically increasing the level of detail of a previously reconstructed scene. This problem is intractable for classic architectures such as vanilla NeRF and I-NGP since the number of layers (resp. grids) of those approaches cannot increase during the reconstruction. Some rare solutions, such as Neural Sparse Voxel Fields \cite{liu2020neural}, combine a pre-trained decoder with a data structure allowing to dynamically allocate voxels to increase the reconstruction resolution. This is however a crucial stake, as being able to adapt the precision of the representation based on the complexity of the scene permits to optimize the computational efficiency (both in speed and memory requirements) with minimal loss in quality.

\section{RING-NeRF}
\label{sec:method}

Rather than focusing on solving one specific problem of NeRF using complex mechanisms, we propose a simple architecture called RING-NeRF constructed with novel inductive biases to tackle NeRF's common issues. 

\subsection{Overview}
\label{sec:overview}

\begin{figure*}[bt]
  \includegraphics[width=\textwidth]{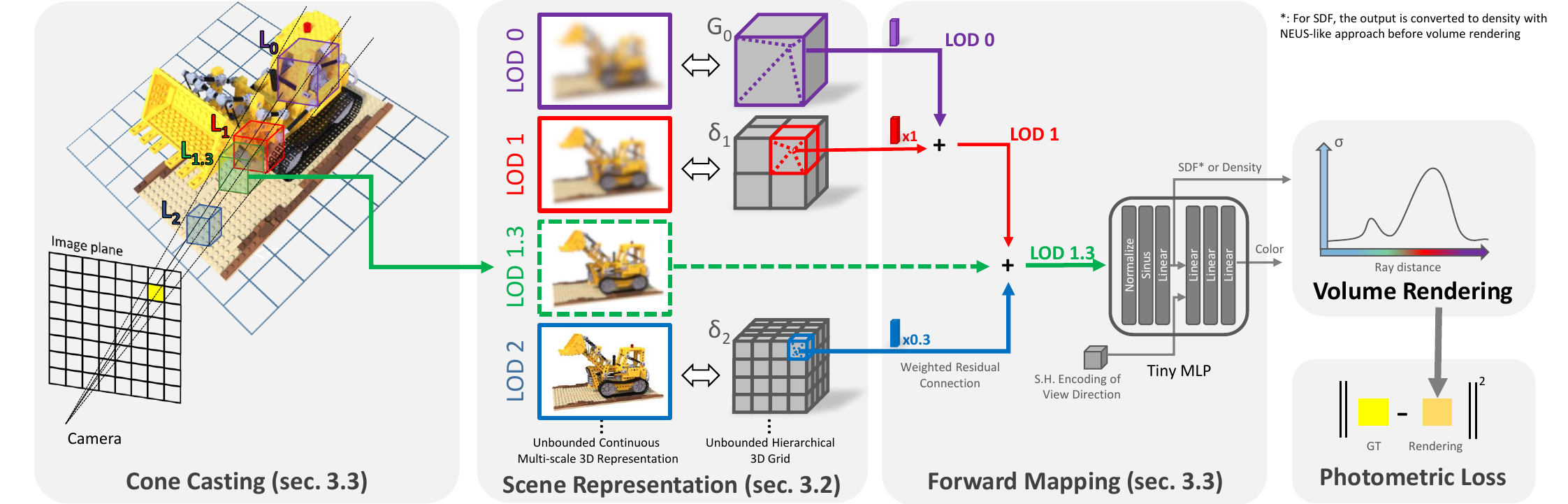}
  \caption{Overview of RING-NeRF: to render a pixel, the casted cone is sampled with cubes. Depending on the cube volume, the corresponding LOD of the scene is selected and the latent feature is computed using a weighted sum of the grid hierarchy. The density (or SDF) and color of the cube are first decoded from the latent feature with a tiny MLP and then integrated with other samples through volume rendering.}
  \label{fig:overview}
\end{figure*}

RING-NeRF relies on the classic NeRF \cite{mildenhall2021nerf} inverse rendering pipeline which is used to reconstruct a 3D scene from a set of localized frames. For a given image pixel (with its camera's pose), a 3D ray is cast and the 3D scene representation is sampled at $N$ various locations along the ray. The resulting density $\sigma_i$ (or SDF converted to density \cite{wang2021neus}) and color $c_i$ of the samples are then combined with usual volume rendering techniques: $\hat{C}(r) = \sum_{i=0}^{N-1} T_i(1-exp(-\sigma_i \delta_i)) c_i$ with $T_i = exp(\sum_{j=0}^{i-1}\sigma_j \delta_j)$ and $\delta_i$ is the distance between samples. The parameters of the scene representation are then optimized by minimizing the  MSE loss $||\hat{C}(r) - C(r)||^2 $ between the rendered ray color and the ground truth pixel value, the rendering being differentiable\footnote{With SDF, an additional Eikonal loss is also being optimized for SDF regularization.}.

The originality of RING-NeRF relies on its neural architecture, illustrated in figure \ref{fig:overview}, which is designed to represent the scene with a continuous and unbounded level of detail without the need for LOD-specific supervision (see section \ref{sec:archi}). We then use this LOD inductive bias to adjust the LOD of the samples with respect to the distance to the camera in contracted space for more accurate renderings (see section \ref{sec:dist_aware_map}). Finally, combined with a continuous coarse-to-fine optimization, RING-NeRF results in a more robust reconstruction process with an intrinsic LOD extensibility property (see section \ref{sec:recons}). 
\subsection{NeRF Representation with Inductive Biases}
\label{sec:archi}
%
Classic grid-based representations, such as the one introduced by I-NGP\cite{muller2022instant}, rely on two key elements: a unique MLP decoder which transforms a latent space into an output space (color, density or SDF), and a 3D (hierarchical) grid of latent features that implicitly defines a unique mapping function from the scene space onto the decoder latent space. 
We can distinguish two different approaches used to extend the mono-scale representation to a multi-scale representation:
conditioning with scale information the decoding process \cite{barron2023zip,turki2023pynerf} and/or defining separate per-LOD mapping functions \cite{turki2023pynerf}.
In both cases, the training  becomes more complex, leading to potential convergence issues as it relies solely on additional specific supervision.  
Instead, we propose to condition the mapping function itself with scale information and introduce inductive biases to guide the convergence. 
\\ \textbf{Scene representation as multi-scale mapping function.}
As illustrated in Figure \ref{fig:overview}, the mapping function at level $N$ is controlled by a grid $G_N$ that is implicitly defined through a recursive refinement process over scale: 


\begin{equation}
\label{eq:reccursiveGrid}
\begin{cases}
G_{0} \leftarrow \delta_{0}\\
G_{N} \leftarrow S(G_{N-1})+\delta_{N}
\end{cases}
\end{equation}
where $\delta_i$ is the deviation defined by the i-th latent feature grid and S is the subdivision scheme consisting in increasing the resolution of $G_{N-1}$ up to the $N$-th grid resolution (with tri-linear interpolation). In practice, the latent vector associated with a point in the scene for a level of detail $N$ is simply obtained by interpolating linearly the point in the cell for each level inferior or equal to $N$, and summing the results. Since linear interpolation is differentiable, our mapping function is also optimizable. The summing of the interpolation is comparable to a residual connection and gives the name of the architecture RING-NeRF for Residual Implicit Neural Grids. 

\begin{figure*}[bt]
  \includegraphics[width=\textwidth]{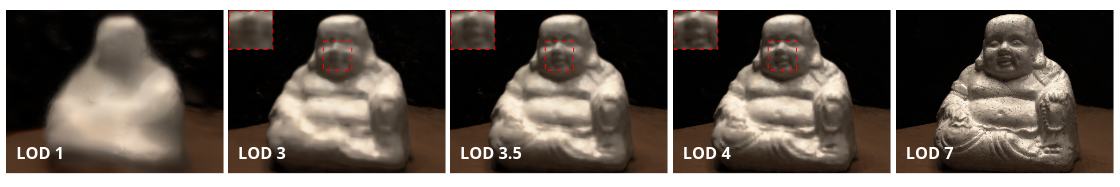}
  \caption{We demonstrate the LOD inductive bias by training our model with a hierarchy of $N=7$ levels where only the last level of the mapping function $G_N$ is supervised. We then compute renders at different levels $L \leq N $. Other examples (including of entire scenes) can be found in the supplementary materials.}
  \label{fig:nat_lod}
\end{figure*}

Such representation provides several advantages. First, since the mapping function is constructed in a top-down manner, refining recursively one level of detail from the previous one, its number of LOD is unbounded and adding a finer level of detail keeps the coarser ones valid. Secondly, because each $\delta_i$ only represents a deviation of the coarser mapping value $G_{i-1}$, a continuous LOD representation can be easily obtained by incorporating a weighting factor $\alpha \in [0,1]$ in the recursive process: 
\begin{equation}
G_{L} \leftarrow S(G_{N-1}) + \alpha \delta_{N}  \label{eq:continuousLOD}  
\end{equation}
with $L \in  [N-1,N ] $ and $ \alpha = L - (N-1)$.
\\ \textbf{Spatial and scale invariance of decoder.} Since our architecture does not rely on a decoder conditioned on position (unlike \cite{takikawa2021neural}) nor scale (unlike \cite{barron2021mip,barron2022mip,barron2023zip,turki2023pynerf}), the decoder latent space is invariant to translations and scale changes in the scene coordinate. 
This property makes RING-NeRF more suited for incremental 
reconstruction in both spatial and scale space, since it ensures that local updates in the spatial and scale domain of the scene can be achieved through the hierarchical grid. The decoder architecture is illustrated in figure \ref{fig:overview} and further details can be found in the supplementary materials. 
\\ \textbf{LOD inductive bias.}
During the reconstruction process, the gradients of the ray samples are backpropagated through the residual connections and aggregated for each grid level.  Due to the pyramidal resolution of the hierarchical grid, a grid code at coarse levels influences a large scene volume and is thus supervised by more ray samples. Therefore, the gradient of a grid code increases as the level's resolution decreases, and as long as the associated samples' gradients are uniform. However, once the backpropagation through the residual connections reach a level whose associated samples have divergent enough gradients (meaning the error is finer than this level's resolution), the result of the aggregation will be mitigated. Hence, the level with the maximum correction is always the coarser level where the samples' gradients are still uniform. This property naturally induces corrections at the proper grid level and LOD. In figure \ref{fig:nat_lod}, we illustrate this inductive bias by displaying different continuous LOD of a scene while the reconstruction is only supervised for its finest LOD. Not only this is a useful property when an unsupervised multi-scale representation is needed (as demonstrated in section \ref{sec:nvs}), this also guarantees a more robust convergence (as illustrated in section \ref{sec:few} and section \ref{sec:sdf}).

\subsection{Distance-Aware Forward Mapping}
\label{sec:dist_aware_map}

\textbf{Cone Casting in contracted space.} In order to accommodate the scene rendering process to the variation of observation distances, we introduce a distance-aware forward mapping mechanism. Similarly to the cone casting of \cite{hu2023tri}, it relies on assigning to each sample a latent feature whose LOD is inversely proportional to the sample-camera distance. However, unlike \cite{hu2023tri,turki2023pynerf}, our model relies on the use of space contraction to allow the reconstruction of unbounded scenes. 

To define the LOD of a sample at distance $d$ of the camera, we first compute the associated volume of the cone cast pixel cube in the world space coordinates (see fig. \ref{fig:overview}). Assuming the pixel is a square of size $c$ at distance $1$ of the camera ($c$ depends on the image resolution and camera's FOV), the volume is thus $V=(dc)^3$. 
To take into account the space contraction, we then proceed to contract the volume. Denoting $J$ the Jacobian of the contraction function at the sample's location $p$, $V_{contract} = V det(J(p))$. In practice, we compute the analytical derivation of the contraction function depending on $p$. More details can be found in the supplementary materials.


Assuming the $N$-th feature grid's resolution in our hierarchy can be written as $f^Nb$ with $b$ the resolution of the grid $\delta_0$ and $f$ the growth factor, finding the appropriate LOD $L \in \mathbb{R}^+$ means finding a virtual grid\footnote{A virtual grid of LOD $L \in \mathbb{R}^+$ corresponds to a grid of resolution $f^Lb$ that is not explicitly stored in memory but whose elements can be computed from other grids.} of resolution $f^Lb$ whose cell's volume is equal to the previously computed volume (in contracted space). Because we are working in the contracted space of size $1$, the volume of a cell of the virtual grid of LOD $L$ is $(\frac{1}{f^L b})^3$. The LOD $L$ associated to one sample in the contracted space is thus given by:
\begin{equation}
(dc)^{3} \operatorname{det}(J(p)) = \left(\frac{1}{f^L b}\right)^3 \iff  L = - \frac{\log\left(dcb \sqrt[3]{\operatorname{det}(J(p))}\right)}{\log(f)}  
\label{eq:da_first}
\end{equation}
Note that this process is close to ZipNeRF \cite{barron2023zip} and VR-NeRF \cite{xu2023VR-NeRF}. However, the first one rather derives a contracted scale factor from its Gaussian samples while the latter directly  computes  the  LOD  in  the  contracted space, which can be considered an approximation of our computation.
\\ \textbf{Forward mapping.} As illustrated in Figure \ref{fig:overview}, we use the determined LOD to compute a distance-dependent weighted sum of the features, which is fed to our decoder and transformed into density (or SDF) and radiance.

\subsection{Continuous Coarse-To-Fine and Resolution Extensibility}
\label{sec:recons}

Recent works proposed to use coarse-to-fine optimization to improve the stability of NeRF models, especially when facing more challenging setups, including with few images \cite{yang2023freenerf} and surface-based models \cite{li2023neuralangelo}. It consists of optimizing progressively the different LOD of the representation, from coarse levels to the most precise ones. The goal of this progressive optimization is to avoid the shape-radiance ambiguity \cite{zhang2020nerf++}  by introducing a strong regularization through LOD restriction, then relaxing progressively this regularization once the coarse geometry of the scene is reconstructed to recover the details of the scene.

The coarse-to-fine reconstruction process of RING-NeRF consists of estimating progressively the LOD of the mapping function from the coarsest to the finest ones. In practice, it implies, during the cone casting, to clamp the samples' LOD up to a maximal LOD $l$, the grids of level $l$ and above being set to zero and not optimized. Moreover, since our architecture provides continuous LOD, the coarse-to-fine optimization can be achieved continuously in the LOD space by using a linear scheduler $l = (l_0 + \frac{n}{n_{ctf}}) \in \mathbb{R}^+$ with $n$ the current epoch, $n_{ctf}$ a hyperparameter describing the speed of the process and $l_0$ defining the number of used grids at initialization; up to a specified maximum resolution. 

Furthermore, the RING-NeRF architecture is more adapted than I-NGP-based architectures \cite{li2023neuralangelo} for coarse-to-fine training. Indeed, for solutions based on the concatenation of features, keeping grid values to zero implies that some dimensions of the decoder's latent space are not supervised. When a grid starts to be optimized, those unsupervised dimensions are suddenly used. The weights of the decoder thus need to be refined, with a global effect on the whole scene. On the opposite, our solution keeps supervising all the dimensions of the decoder latent space, and when a new grid gets optimized, it only implies more degrees of freedom to define the mapping function between the scene space and the decoder latent space.
This also means that our scene reconstruction can be refined by adding dynamically new grid levels without modifying the decoder's weights or previously trained grids, as we demonstrate experimentally in section \ref{sec:exp_extensibility}. This resolution extensibility property opens the path to adaptive resolution models, where the precision used to describe an area depends on the details needed, to optimize efficiency both in memory consumption and training duration.

\section{Experiments}
\label{sec:exps}

In these experiments, we intend to highlight the versatility of RING-NeRF by evaluating it on several tasks. After introducing implementation details in section \ref{sec:imp}, we evaluate our model on novel view synthesis with changes in observation distances (sec. \ref{sec:nvs}). Then, we explore how robust is RING-NeRF first with few view reconstruction (sec. \ref{sec:few}) and then without scene-specific initialization for SDF reconstruction (sec. \ref{sec:sdf}). Finally, we demonstrate the capacity of our architecture to perform LOD extensibility, as a first step towards adaptive reconstruction (sec. \ref{sec:exp_extensibility}).

\subsection{Implementation}
\label{sec:imp}
Our model is based on the PyTorch framework \textit{Nerfstudio} \cite{tancik2023nerfstudio}. We build upon its core method named Nerfacto, which combines ideas from several papers for fast and qualitative renders of unbounded complex scenes. This makes it an accessible baseline with a state-of-the-art quality/time ratio.  
Because NeRF pipelines contain a high number of small but decisive choices of implementation (eg. some frameworks choose to train their models image by image while Nerfstudio jointly and randomly samples across all images), we decided to use as much as possible Nerfstudio-based baselines for fairer comparisons. All of these models are trained on one Nvidia-A100 GPU. The reported times correspond to the approximated training duration of the models. Configuration details, further experiments and ablatives on our contributions can be found in the supplementary materials.

\begin{figure*}[b]
\includegraphics[width=\textwidth]{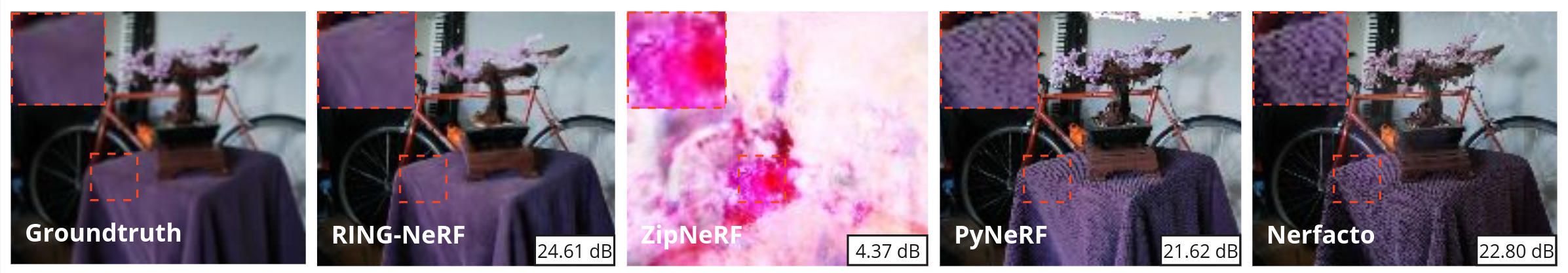}%
}{%
  \caption{Examples of image renderings at 1/8th resolution from models solely trained with the full resolution images. RING-NeRF is the only method capable of producing coherent aliasing-free renderings thanks to its LOD inductive bias.} 
\label{fig:nat_aliasing}
\end{figure*}

%
%

\subsection{Novel View Synthesis and Anti-Aliasing}
\label{sec:nvs}


This experiment aims to evaluate the reconstruction quality through the ability to synthesize viewpoints that are not supervised during the reconstruction. These new viewpoints differ from the angle of observation, but also from the distance of observation. The latter is particularly important since a reconstruction or a rendering process that does not take correctly into account the distance of observation leads to artefacts, from over-contrasted rendering to aliasing phenomenon. The challenge is to avoid these artefacts while keeping the reconstruction and rendering process as fast as possible.
\\ \textbf{Dataset.} The evaluation relies on the dataset introduced by Mip-NeRF-360 \cite{barron2022mip}. This dataset is composed of 9 scenes, each containing both a central area and complex background in both inside and outside setups. Since the trajectory keeps a constant distance to the central part, each viewpoint is represented with a pyramid of 4 different image resolutions to simulate a variation of the distance of observation combined with a change of the camera FOV, following the Mip-NeRF \cite{barron2021mip} original anti-aliasing evaluation pipeline. 
\\ \textbf{Algorithms.}  We compare our solution against several grid-based NeRF baselines, both with (PyNeRF \cite{turki2023pynerf} and Zip-NeRF \cite{barron2023zip}) and without anti-aliasing processing (Nerfacto), using their Nerfstudio implementations. 
\\ \textbf{Protocol.} 
 For each scene, we train the models with two different setups: the \textit{mono-scale} setup that uses only the image with the highest resolution for each viewpoint, and the \textit{multi-scale} setup which uses the whole pyramid of images
for each viewpoint. Note that, for these two setups, we evaluate the performances on the whole resolution pyramid, with usual metrics (PSNR, SSIM, LPIPS).
\begin{table}[tb]
\centering
\caption{Novel View Synthesis performances for the \textbf{Mono-Scale setup} (trained on the full resolution images only) on the 360 Dataset. The indicated resolutions refer to the resolution of the renders.}
\resizebox{\textwidth}{!}{%

\begin{tabular}{l|lll||lll||lll||lll|l}
   & \multicolumn{3}{c||}{Full Res.}                                                                                   & \multicolumn{3}{c||}{1/2 Res.}                                                                                   & \multicolumn{3}{c||}{1/4 Res.}  & \multicolumn{3}{c||}{1/8 Res.} &                                                                             \\
     & \multicolumn{1}{c|}{PSNR ↑} & \multicolumn{1}{c|}{SSIM ↑} & \multicolumn{1}{c||}{LPIPS ↓} & \multicolumn{1}{c|}{PSNR ↑}  & \multicolumn{1}{c|}{SSIM ↑} & \multicolumn{1}{c||}{LPIPS ↓} & \multicolumn{1}{c|}{PSNR ↑} & \multicolumn{1}{c|}{SSIM ↑} &\multicolumn{1}{c||}{LPIPS ↓} & \multicolumn{1}{c|}{PSNR ↑} & \multicolumn{1}{c|}{SSIM ↑} & \multicolumn{1}{c||}{ LPIPS ↓}  & Time ↓    \\ \hline
     Nerfacto \cite{tancik2023nerfstudio} & \multicolumn{1}{c|}{ 27.09}   & \multicolumn{1}{c|}{ 0.779}   &\multicolumn{1}{c||}{ 0.181}   &     \multicolumn{1}{c|}{\cellcolor{orange!50} 26.81}   &\multicolumn{1}{c|}{\cellcolor{orange!50} 0.782}   &\multicolumn{1}{c||}{\cellcolor{orange!50} 0.162}   & \multicolumn{1}{c|}{\cellcolor{orange!50} 25.22}   &\multicolumn{1}{c|}{\cellcolor{orange!50} 0.711}   &\multicolumn{1}{c||}{\cellcolor{orange!50} 0.234}   & \multicolumn{1}{c|}{\cellcolor{orange!50} 23.32}   &     \multicolumn{1}{c|}{\cellcolor{orange!50} 0.636}   &        \multicolumn{1}{c||}{\cellcolor{orange!50} 0.297}    &  \multicolumn{1}{c}{\cellcolor{red!50} 0.45h }\\ 
PyNeRF \cite{turki2023pynerf} & \multicolumn{1}{c|}{\cellcolor{yellow!50}27.87}   & \multicolumn{1}{c|}{\cellcolor{orange!50}0.802}   & \multicolumn{1}{c||}{\cellcolor{yellow!50} 0.160}   & \multicolumn{1}{c|}{\cellcolor{yellow!50} 21.93}   & \multicolumn{1}{c|}{\cellcolor{yellow!50}0.713}   & \multicolumn{1}{c||}{\cellcolor{yellow!50}0.211}   &  \multicolumn{1}{c|}{\cellcolor{yellow!50}21.15}   & \multicolumn{1}{c|}{\cellcolor{yellow!50}0.662}   & \multicolumn{1}{c||}{\cellcolor{yellow!50}0.264}   &  \multicolumn{1}{c|}{\cellcolor{yellow!50}20.39}       &     \multicolumn{1}{c|}{\cellcolor{yellow!50}0.619}      &     \multicolumn{1}{c||}{\cellcolor{yellow!50}0.312}    &   \multicolumn{1}{c}{\cellcolor{yellow!50} 0.96h}\\
Zip-NeRF \cite{barron2023zip}  & \multicolumn{1}{c|}{\cellcolor{orange!50} 28.06}   & \multicolumn{1}{c|}{\cellcolor{red!50} 0.808}   & \multicolumn{1}{c||}{\cellcolor{red!50} 0.154}   &                  \multicolumn{1}{c|}{ 16.58}   & \multicolumn{1}{c|}{0.596}   & \multicolumn{1}{c||}{ 0.319}   & \multicolumn{1}{c|}{ 11.88}   & \multicolumn{1}{c|}{ 0.424}   &  \multicolumn{1}{c||}{ 0.465}   & \multicolumn{1}{c|}{9.66}      &    \multicolumn{1}{c|}{0.323}     &    \multicolumn{1}{c||}{ 0.523}   & \multicolumn{1}{c}{ 1.10h}  \\  \hline
RING-NeRF & \multicolumn{1}{c|}{\cellcolor{red!50} 28.09}   & \multicolumn{1}{c|}{\cellcolor{yellow!50} 0.799}   & \multicolumn{1}{c||}{\cellcolor{orange!50} 0.157}   & \multicolumn{1}{c|}{\cellcolor{red!50} 27.18}   & \multicolumn{1}{c|}{\cellcolor{red!50} 0.804}   & \multicolumn{1}{c||}{\cellcolor{red!50} 0.138}   & \multicolumn{1}{c|}{\cellcolor{red!50} 25.82}   & \multicolumn{1}{c|}{\cellcolor{red!50} 0.786}   & \multicolumn{1}{c||}{\cellcolor{red!50} 0.142}   & \multicolumn{1}{c|}{\cellcolor{red!50} 24.38}     &    \multicolumn{1}{c|}{\cellcolor{red!50} 0.737}     &    \multicolumn{1}{c||}{\cellcolor{red!50} 0.167}    & \multicolumn{1}{c}{\cellcolor{red!50} 0.45h} \\ 
\end{tabular}
}
\label{tab:nvs}
\end{table}
\begin{table*}[tb]
\centering
\caption{Novel View Synthesis performances for the \textbf{Multi-Scale setup} (trained jointly on every resolution) on the 360 Dataset. The indicated resolutions refer to the resolution of the renders.}
\resizebox{\textwidth}{!}{%
\begin{tabular}{l|lll||lll||lll||lll|l}
     & \multicolumn{3}{c||}{Full Res.}                                                                                   & \multicolumn{3}{c||}{1/2 Res.}                                                                                   & \multicolumn{3}{c||}{1/4 Res.}  & \multicolumn{3}{c||}{1/8 Res.} &                                                                             \\
       & \multicolumn{1}{c|}{PSNR ↑} & \multicolumn{1}{c|}{SSIM ↑} & \multicolumn{1}{c||}{LPIPS ↓} & \multicolumn{1}{c|}{PSNR ↑}  & \multicolumn{1}{c|}{SSIM ↑} & \multicolumn{1}{c||}{LPIPS ↓} & \multicolumn{1}{c|}{PSNR ↑} & \multicolumn{1}{c|}{SSIM ↑} &\multicolumn{1}{c||}{LPIPS ↓} & \multicolumn{1}{c|}{PSNR ↑} & \multicolumn{1}{c|}{SSIM ↑} & \multicolumn{1}{c||}{ LPIPS ↓}  & Time ↓    \\ \hline
     Nerfacto \cite{tancik2023nerfstudio} & \multicolumn{1}{c|}{ 25.98}   & \multicolumn{1}{c|}{0.668}   &\multicolumn{1}{c||}{ 0.367}   &     \multicolumn{1}{c|}{ 27.31}   &\multicolumn{1}{c|}{ 0.792}   &\multicolumn{1}{c||}{ 0.176}   & \multicolumn{1}{c|}{ 27.16}   &\multicolumn{1}{c|}{ 0.805}   &\multicolumn{1}{c||}{ 0.139}   & \multicolumn{1}{c|}{ 25.30}   &     \multicolumn{1}{c|}{ 0.743}   &        \multicolumn{1}{c||}{ 0.199}    &  \multicolumn{1}{c}{\cellcolor{red!50} 0.45h }\\ 
PyNeRF \cite{turki2023pynerf} & \multicolumn{1}{c|}{\cellcolor{orange!50}27.65}   & \multicolumn{1}{c|}{\cellcolor{orange!50}0.781}   & \multicolumn{1}{c||}{\cellcolor{orange!50} 0.194}   & \multicolumn{1}{c|}{\cellcolor{orange!50} 29.47}   & \multicolumn{1}{c|}{\cellcolor{orange!50}0.857}   & \multicolumn{1}{c||}{\cellcolor{red!50}0.092}   &  \multicolumn{1}{c|}{\cellcolor{red!50}30.51}   & \multicolumn{1}{c|}{\cellcolor{yellow!50}0.887}   & \multicolumn{1}{c||}{\cellcolor{orange!50}0.063}   &  \multicolumn{1}{c|}{\cellcolor{orange!50}30.86}       &     \multicolumn{1}{c|}{\cellcolor{yellow!50}0.893}      &     \multicolumn{1}{c||}{\cellcolor{yellow!50}0.055}    &   \multicolumn{1}{c}{\cellcolor{yellow!50} 0.96h}\\
Zip-NeRF \cite{barron2023zip}  & \multicolumn{1}{c|}{\cellcolor{yellow!50} 27.54}   & \multicolumn{1}{c|}{\cellcolor{orange!50} 0.781}   & \multicolumn{1}{c||}{\cellcolor{red!50} 0.191}   &                  \multicolumn{1}{c|}{ \cellcolor{yellow!50}  29.34}   & \multicolumn{1}{c|}{\cellcolor{red!50} 0.858}   & \multicolumn{1}{c||}{\cellcolor{yellow!50}  0.099}   & \multicolumn{1}{c|}{ \cellcolor{yellow!50}  30.36}   & \multicolumn{1}{c|}{ \cellcolor{red!50} 0.889}   &  \multicolumn{1}{c||}{\cellcolor{yellow!50}  0.064}   & \multicolumn{1}{c|}{\cellcolor{red!50} 30.90}      &    \multicolumn{1}{c|}{\cellcolor{orange!50}  0.900}     &    \multicolumn{1}{c||}{ \cellcolor{red!50}  0.047}   & \multicolumn{1}{c}{ 1.10h}  \\  \hline
RING-NeRF & \multicolumn{1}{c|}{\cellcolor{red!50} 27.66}   & \multicolumn{1}{c|}{\cellcolor{red!50} 0.782}   & \multicolumn{1}{c||}{\cellcolor{orange!50} 0.194}   & \multicolumn{1}{c|}{\cellcolor{red!50} 29.55}   & \multicolumn{1}{c|}{\cellcolor{yellow!50} 0.856}   & \multicolumn{1}{c||}{\cellcolor{orange!50} 0.094}   & \multicolumn{1}{c|}{\cellcolor{red!50} 30.51}   & \multicolumn{1}{c|}{\cellcolor{orange!50} 0.888}   & \multicolumn{1}{c||}{\cellcolor{red!50} 0.062}   & \multicolumn{1}{c|}{\cellcolor{orange!50} 30.86}     &    \multicolumn{1}{c|}{\cellcolor{red!50} 0.901}     &    \multicolumn{1}{c||}{\cellcolor{orange!50} 0.048}    & \multicolumn{1}{c}{\cellcolor{red!50} 0.45h} \\ 
\end{tabular}
}

\label{tab:AA}
\end{table*}
%
%
%
%
\\ \textbf{Results.} First of all, regarding novel view synthesis quality using the mono-scale setup for both training and testing, as referred to in the "Full Res." column of table \ref{tab:nvs}, we notice that our architecture provides on par performances with Zip-NeRF, slightly better results than PyNeRF, and a more important gap with Nerfacto. While simple, RING-NeRF succeeds in performing state-of-the-art performances on a real single-scale dataset.
Regarding quality for the multi-scale setup, as presented in table \ref{tab:AA}, we observe that all the algorithms that consider the distance of observation perform very similarly in terms of quality.
On the other side, Nerfacto performs quite poorly. Its performances are especially low at the coarsest resolution, with overflowing artefacts, as illustrated in figure \ref{fig:results}. 

Finally, we evaluate the capacity of the models to generalize over new resolutions by training them on full-resolution images and then evaluate them on smaller resolutions. Qualitative results are shown in figure \ref{fig:nat_aliasing} while quantitative results can be found in the 1/2-th, 1/4-th and 1/8-th res. columns of table \ref{tab:nvs}. Because most anti-aliasing methods need LOD-specific supervision to correctly function (including Zip-NeRF and PyNeRF), they cannot render coherent anti-aliased images in this setup. However, PyNeRF behaves better than ZipNeRF with coherent although very aliased renderings, as it decides to limit the LOD used for rendering inside the range of LOD seen during training. RING-NeRF, with his LOD inductive bias, is the only architecture capable of producing anti-aliased renderings from novel observation distances and thus outperforms every other method. 
While this experiment can seem somewhat esoteric as training on a multi-resolution images pyramid is rather easy, this increases GPU memory and total training time. Moreover, depending on the resolution and the trajectory, it is not trivial to choose the accurate number of scales in the image pyramid to supervise correctly every grid in the hierarchy and especially the coarsest grids.

Regarding the reconstruction processing time reported in the "Time" column of both table \ref{tab:nvs} and table \ref{tab:AA}, since RING-NeRF does not rely on the multiplication of either sample or decoder, it processes as quickly as the fastest Nerfacto both on mono-scale and multi-scale setups. On the other hand, the other anti-aliasing methods, PyNeRF with its per-LOD MLP and Zip-NeRF with its super-sampling, are approximately 2.5 times slower. 

In conclusion, our solution provides the best quality-speed trade-off since it is both on par with the best quality method and the fastest method, in mono-scale as well as in multi-scale setups. Furthermore, RING-NeRF is the only solution capable of creating coherent and anti-aliased renderings when facing novel observation distances unseen during training.
\subsection{Few Viewpoints Supervision}
\label{sec:few}



\begin{table}[b!]
\centering
\caption{Performances of reconstruction from few viewpoints on the DTU dataset. The reported metrics are computed based on the mask of the object.} 
\resizebox{0.9\columnwidth}{!}{%
\begin{tabular}{c|c c c| c c c|c c c|c}
                   & \multicolumn{3}{c|}{PSNR ↑} & \multicolumn{3}{c|}{SSIM ↑}  &\multicolumn{3}{c|}{LPIPS ↓}   &       \\ 
                  \#Images & 3 & 6 &9 &3 &6 &9 &3 &6 &9 & Time ↓ \\\hline
Mip-NeRF \cite{barron2021mip} & 8.68 & 16.54 & \cellcolor{orange!50} 23.58 & 0.571 & 0.741 & \cellcolor{orange!50}  0.879 &  0.353 & 0.198 &0.092 & 2.56h \\
FreeNeRF \cite{yang2023freenerf}          &  \cellcolor{red!50} 19.92 & \cellcolor{red!50} 23.25 &\cellcolor{red!50}  25.38 & \cellcolor{red!50} 0.787 & \cellcolor{red!50} 0.844 & \cellcolor{red!50} 0.888  & \cellcolor{red!50} 0.135 & \cellcolor{red!50} 0.095 & \cellcolor{red!50} 0.067  & 2.56h   \\
\hline
Nerfacto   \cite{tancik2023nerfstudio}       & 9.35 & 9.75 & 9.78  & 0.567 & 0.604 & 0.647 & 0.385 & 0.331 & 0.326   & \cellcolor{red!50}0.15h \\ 
Nerfacto+          & 13.61 & 16.61 & 19.33 & 0.639 & 0.699 &0.759 & 0.276 & 0.218 & 0.151 & \cellcolor{red!50}0.15h \\ 

 \hline
RING-NeRF          & \cellcolor{orange!50} 16.18 & \cellcolor{orange!50} 20.47 & \cellcolor{yellow!50} 23.19 & \cellcolor{orange!50} 0.713 & \cellcolor{orange!50} 0.808 & \cellcolor{yellow!50} 0.847 & \cellcolor{orange!50} 0.200 & \cellcolor{orange!50} 0.127 & \cellcolor{orange!50} 0.085  & \cellcolor{red!50}0.15h  \\ 
 w/ discrete CtF & \cellcolor{yellow!50} 15.79 &  \cellcolor{yellow!50} 20.16 &  22.93  & \cellcolor{yellow!50} 0.706 &  \cellcolor{yellow!50} 0.785 &  \cellcolor{yellow!50} 0.847 & \cellcolor{yellow!50} 0.201 &  \cellcolor{orange!50} 0.127 &  \cellcolor{orange!50} 0.085 & \cellcolor{red!50}0.15h\\ 

\end{tabular}%
}

\label{tab:fewViews}
\end{table}

This experiment aims to evaluate the influence of the RING-NERF architecture and pipeline on the reconstruction robustness to limited supervision viewpoints. 
%
\\ \textbf{Dataset.} We evaluate our contribution on the object-centric real dataset DTU \cite{jensen2014large}, often used in few-viewpoints evaluations.
\\ \textbf{Algorithms.} 
We compare our architecture against several baselines: Mip-NeRF \cite{barron2021mip}, and FreeNerf \cite{yang2023freenerf} (a state-of-the-art method for this task), for vanilla architectures and Nerfacto for grid-based architectures.  To better demonstrate the intrinsic stability brought by RING-NeRF, we also developed the Nerfacto+ architecture, which corresponds to a Nerfacto architecture coupled with a coarse-to-fine training based on a progressive activation of the grids (the rest of the decoder's input being filled with zeros).  For Nerfacto+ and RING-NeRF, we also add FreeNeRF's loss that penalizes the density of the first $M=10$ samples of each ray to reduce as much as possible artefacts in front of the cameras. As an ablative experiment, we also evaluate RING-NeRF using discrete coarse-to-fine (fixed LOD increment of 1 rather than the proposed continuous increase). 
%
\\ \textbf{Protocol.} We follow FreeNeRF's evaluation pipeline, including the number of supervision viewpoints (3 to 9), the choice of these views among the dataset and the evaluations using masks of the object. Since we are using the same protocol, the results of FreeNeRF and Mip-NeRF were taken out of FreeNeRF's article. 
%
\\ \textbf{Results.}
Evaluation results are reported in table \ref{tab:fewViews}. We first notice an important difference between vanilla and grid-based baselines. While Mip-NeRF faces troubles in reconstructing the scene with 3 views, the method seems to find coherency when adding more images. However, Nerfacto struggles much more to form a consistent 3D scene even when using 9 images (see figure \ref{fig:results}). Even though the grid-based baseline is way faster to train, its design implies more instability when facing a small number of images. 
This does not mean however that few viewpoints are incompatible with grid-based methods. Using progressive training coupled with a very simple and generalizable regularization, both Nerfacto+ and our architecture succeed in creating coherent geometry. Nonetheless, RING-NeRF considerably outperforms Nerfacto+, with a PSNR difference varying from 3 to 4, demonstrating the stability increase brought by our architecture, and also outperforms Mip-NeRF for the configuration with 3 and 6 supervision images. The discrete coarse-to-fine version of RING-NeRF performs in-between Nerfacto+ and the complete RING-NeRF. This showcases both the intrinsic interest of the proposed architecture against the Nerfacto+ and the relevance of the continuous coarse-to-fine mechanism. FreeNerf remains the best performer of all in terms of quality, but with a reconstruction time that is extremely slower than RING-NeRF, the former achieving its reconstruction in 2.56 hours while the latter only requires less than 10 minutes. RING-NeRF thus offers a better quality-speed trade-off for the few-view reconstruction issue (see figure \ref{fig:results}). 
\subsection{SDF Reconstruction without Initialization}
\label{sec:sdf}
SDF reconstruction is known to be a more unstable process than density-based NeRF \cite{atzmon2020sal}, requiring a scene-specific initialization to converge. This initialization becomes an issue in complex environments and incremental setups, where several types of scenes can co-exist and are not necessarily known beforehand. Therefore, in this experiment, we evaluate the ability of RING-NeRF and other SotA architectures to achieve SDF reconstruction without scene-specific initialization.
%
\\ \textbf{Dataset.} The evaluation is achieved on a subset of 7 scenes of the \textit{Replica}\cite{straub2019replica} synthetic indoor dataset. A Tanks \& Temples \cite{knapitsch2017tanks} example is provided in supplementary materials with corresponding analysis.
\\ \textbf{Algorithms.} We compare our architecture to two SDF methods, all of them implemented in the same \textit{SDFStudio} \cite{Yu2022SDFStudio} branch of the \textit{Nerfstudio} framework for fairer comparisons: NeuS-facto, an adaptation of NeuS for grid-based methods with Nerfacto modules, and an implementation of NeuralAngelo.  For these experiments, our model is built upon the NeuS-facto baseline, using in particular the same NeuS-based SDF-to-density transformation\cite{wang2021neus}.
\\ \textbf{Protocol.} 
To evaluate the impact of the architecture and pipeline over the convergence and stability, we suppress the inverted sphere SDF initialization scheme and use a random initialization for the model.
\begin{table}[t]

\centering
\caption{SDF reconstruction performances when foregoing the scene-specific SDF Initialization on the Replica Dataset. The Chamfer distance is in centimeters.} 
\label{tab:sdf}
\resizebox{0.7\columnwidth}{!}{%
\begin{tabular}{c|c c c|c|c}
                 & PSNR ↑ & SSIM ↑ & LPIPS ↓ &  Chamfer-L1 ↓ & Training Time ↓ \\ \hline

NeuS-facto\cite{Yu2022SDFStudio}           &  \cellcolor{yellow!50} 24.62   & \cellcolor{yellow!50}  0.778   & \cellcolor{yellow!50} 0.347  &   \cellcolor{yellow!50} 17.61  & \cellcolor{orange!50} 1.4h  \\  \hline

NeuralAngelo\cite{li2023neuralangelo}   &  \cellcolor{orange!50} 30.79     &   \cellcolor{orange!50}   0.916   &  \cellcolor{orange!50}  0.0761      &  \cellcolor{orange!50}  13.69  & \cellcolor{yellow!50} 3.40 h          \\ 

\hline

RING-NeRF           & \cellcolor{red!50} 37.18   &   \cellcolor{red!50}0.969    &  \cellcolor{red!50}0.0194     &  \cellcolor{red!50} 5.71    & \cellcolor{red!50} 1.28 h   \\

\end{tabular}%
}

\end{table}
\\ \textbf{Results.} Evaluation results are shown in table \ref{tab:sdf}.
First of all, NeuS-facto faces low rendering and reconstruction metrics, due to catastrophic failure in most of the tested scenes (see figure \ref{fig:results}) 
since the model tends to re-draw the 2D images in front of the camera.   
Regarding NeuralAngelo, its relatively high PSNR demonstrates its ability to synthesize satisfying RGB. However, as illustrated in figure \ref{fig:results}  and highlighted by the reconstruction metrics, the underlying geometry of the scene is poorly reconstructed, without any fine details. Finally, our method RING-NeRF is by far the best performer, with much higher PSNR and a better geometry including fine details (see figure \ref{fig:results}), although a bit noisy.
The simple architecture of RING-NeRF also permits faster epochs, thus faster training.

\subsection{LOD Extensibility}
\label{sec:exp_extensibility}





\begin{figure*}[tb]
\centering
  \includegraphics[width=\textwidth]{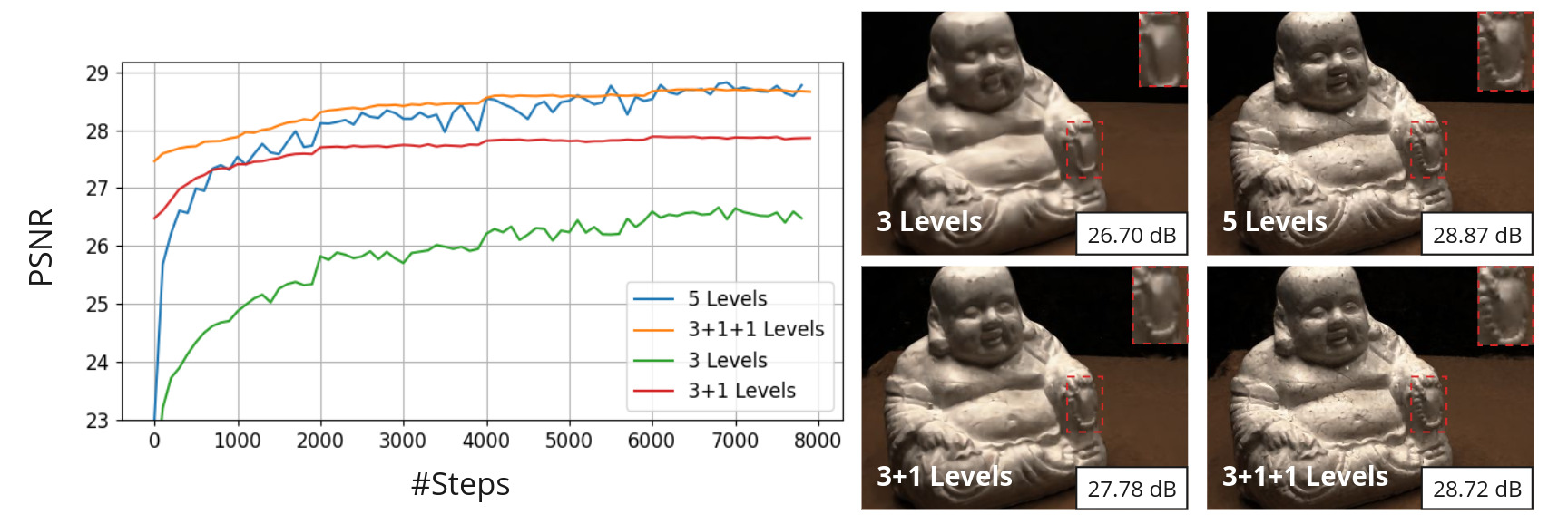}
  \caption{Learning curves and final renderings of RING-NeRF models with different grid configurations trained either jointly or incrementally. }
\label{fig:extend}
\end{figure*}

This experiment aims to demonstrate RING-NeRF's unique ability to increase dynamically the level of detail of the scene representation. 
%
\\ \textbf{Dataset.}  The scan 114 of the DTU dataset is used for this experiment.
\\ \textbf{Algorithms.} Because I-NGP architectures \cite{muller2022instant,barron2023zip,li2023neuralangelo} cannot perform LOD extensibility (due to the fixed decoder's input size), only RING-NeRF is evaluated.
\\ \textbf{Protocol.}  
We train our model using two different configurations : one low resolution with a 3 levels grid hierarchy and one high resolution with 5 levels (the grid resolutions are $16$, $32$, $64$, $128$ and $256$). For these two configurations named "3 levels" and "5 levels", every grid and the decoder are trained simultaneously.
We proceed to showcase the extensibility of RING-NeRF by adding grids to the "3 levels" configuration that is previously trained. We first train one grid of resolutions $128$ ("3+1 levels" in Figure \ref{fig:extend}) with the three initial grids and the decoder frozen and then train another grid of resolutions $256$ ("3+1+1 levels" in Figures \ref{fig:extend}) with the four grids and the decoder frozen.
\\ \textbf{Results.} Figure \ref{fig:extend} shows that the configuration "3+1+1 levels" results in the same rendering quality than the "5 levels" one. This demonstrates the ability of our model to dynamically change the resolution of the grid hierarchy. This is an important step towards the development of an adaptive architecture which locally chooses the resolution of the representation based on the scene's content. This allows to drastically reduces the number of parameters, helping in improving the memory footprint, the training duration and the model’s robustness.

\section{Conclusion and Perspectives}
\label{sec:ccl}

In this work, we introduced RING-NeRF, a simple and versatile NeRF pipeline that provides two inductive biases by design: a continuous multi-scale representation of the scene, and an invariance of the decoder latent space over spatial and scale domains. Coupled with a distance-aware forward mapping and a continuous coarse-to-fine reconstruction process, our pipeline demonstrated experimentally its versatility with on-par performances with dedicated state-of-the-art solutions for anti-aliasing or reconstruction from few viewpoints. It even outperforms them in terms of robustness to scene-specific initialization for SDF reconstruction. Furthermore, it is highly efficient and is not limited to object-centric scenes.

Future work will study the impact of RING-NeRF on other challenging use cases, such as facing inaccurate camera poses \cite{park2023camp} and SLAM \cite{zhu2022nice}. We will also use the extensibility property of our architecture to develop memory-efficient sparse Neural Fields, which is considered to be a limit of most grid-based models.

\section*{Acknowledgements} This publication was made possible by the use of the CEA List FactoryIA supercomputer, financially supported by the Ile-de-France Regional Council.

\bibliographystyle{splncs04}

\clearpage
\setcounter{page}{1}

\interfootnotelinepenalty=10000
\title{Supplementary Materials of RING-NeRF} 



\titlerunning{RING-NeRF}

\author{Doriand Petit\inst{1,2} \and
Steve Bourgeois\inst{1} \and
Dumitru Pavel\inst{1} \and
Vincent Gay-Bellile\inst{1} \and
Florian Chabot\inst{1} \and
Loïc Barthe\inst{2}}

\authorrunning{D. Petit et al.}

\institute{Université Paris-Saclay, CEA, List,
F-91120, Palaiseau, France \and
IRIT, Université Toulouse III, CNRS, France
}
\maketitle



\begin{figure*}[!ht]
    \captionsetup{type=figure}
    \includegraphics[width=\textwidth]{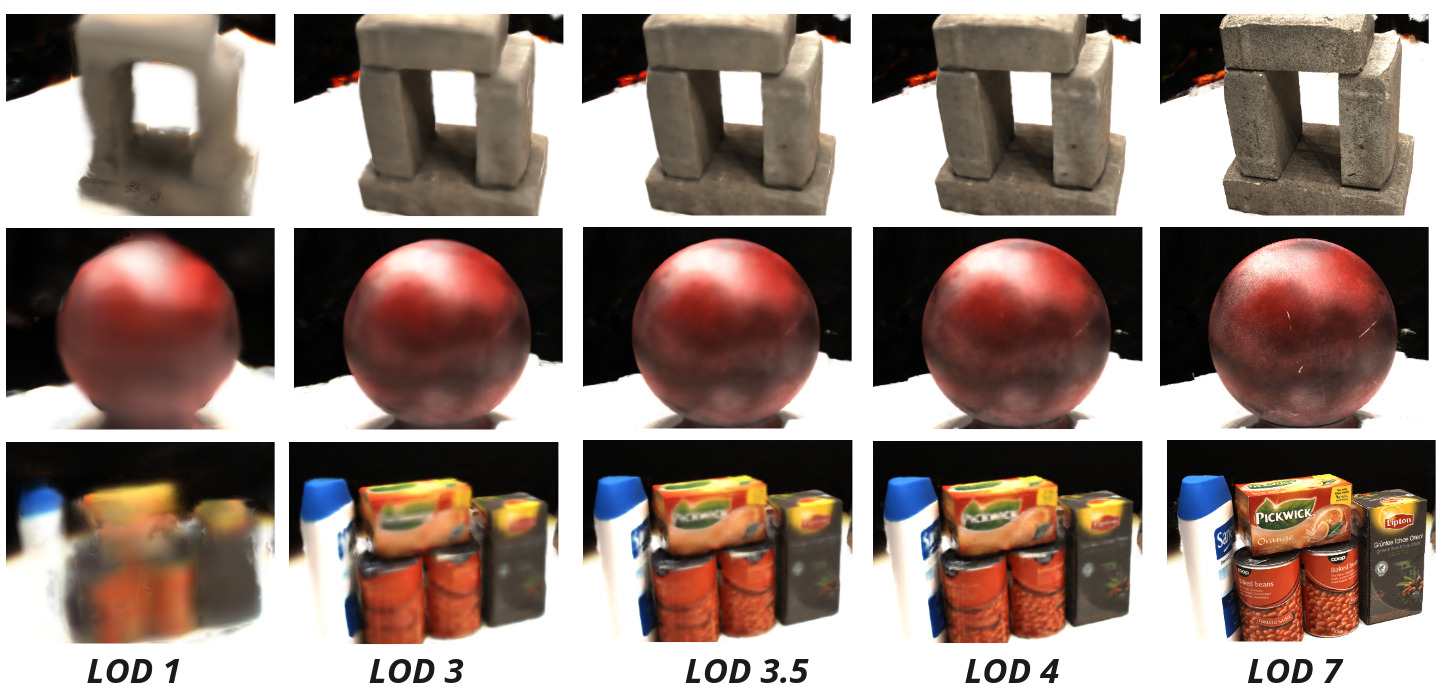}
    \captionof{figure}{Different LODs Outputs of the model when only trained on the last level $L=7$. We observe that, even without supervising intermediate LODs during the training, a  notion  of LOD is captured in the scene reconstruction. We also observe visually continuous LOD since, as expected, the level $L=3.5$ outputs a 3D representation in between LOD $L=3$ and $L=4$ in term of details. }
    \label{fig:lod}
\end{figure*}

\begin{figure*}[!ht]
    \captionsetup{type=figure}
    \includegraphics[width=\textwidth]{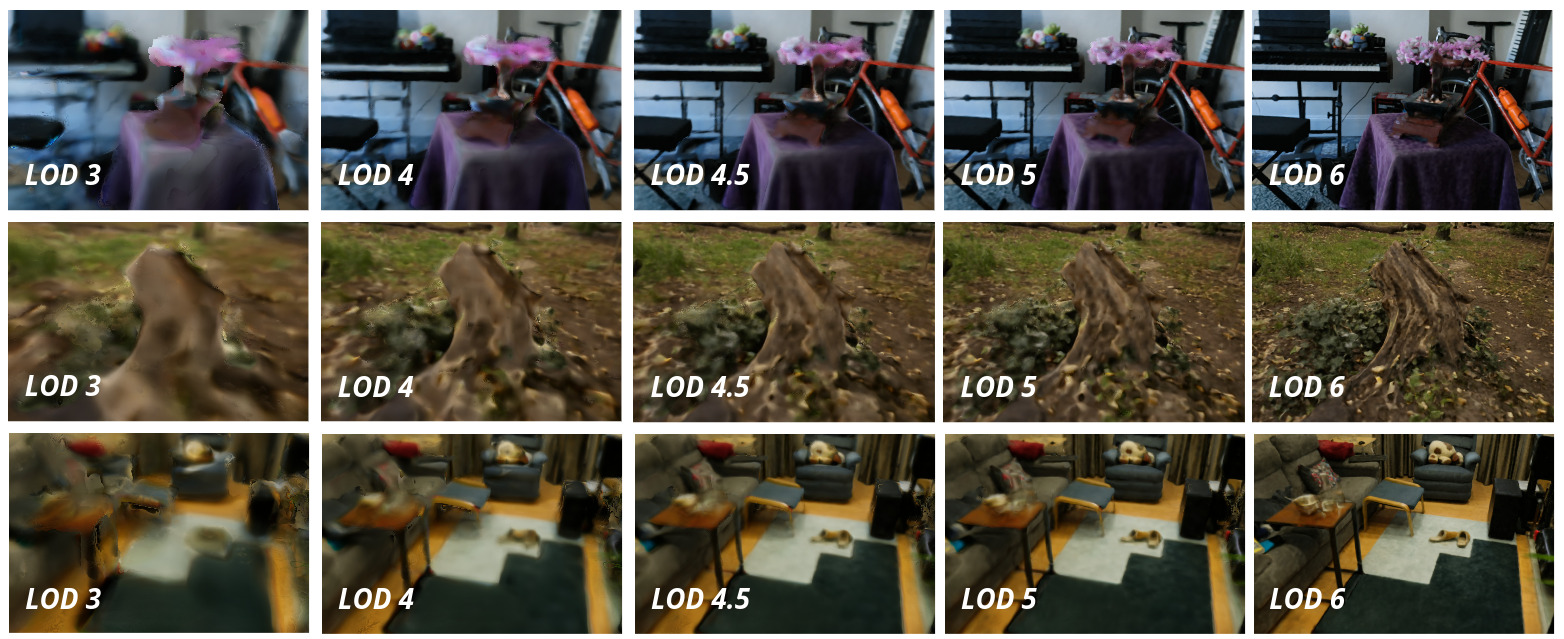}
    \captionof{figure}{Different LODs Outputs of the model when only trained on the last level $L=7$. This illustrates the LOD inductive biases even on more complex scenes than DTU single objects. }
    \label{fig:lod_mip}
\end{figure*}

\begin{figure*}
    \captionsetup{type=figure}
    \includegraphics[width=\textwidth]{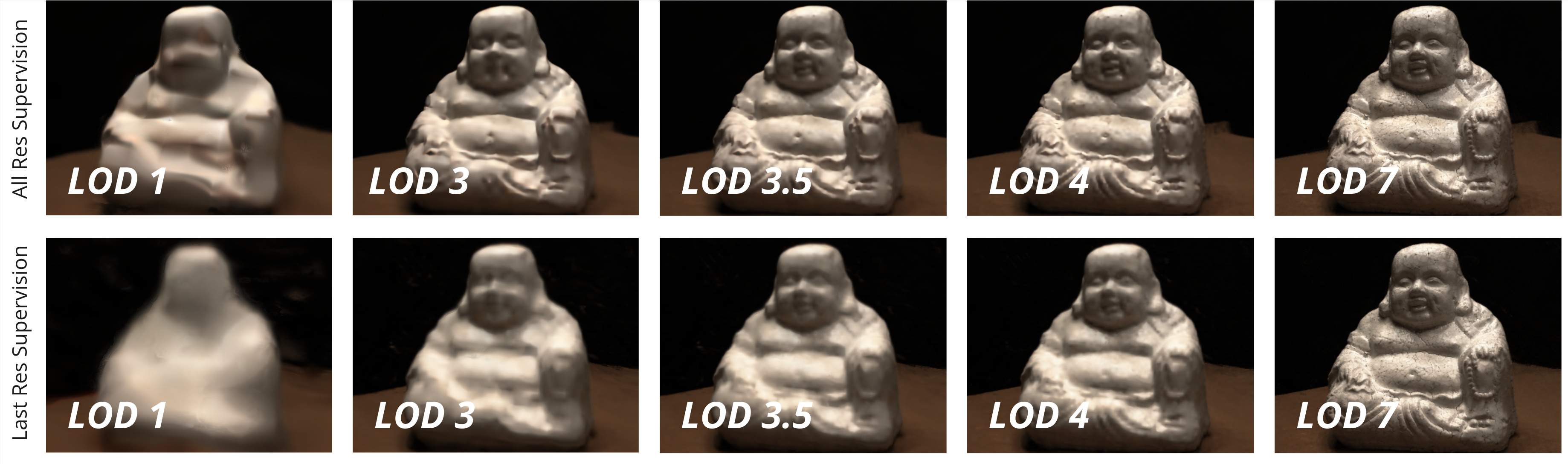}
    \captionof{figure}{Comparison of LOD when supervising (a) every level of detail or (b) solely the finest level of detail}
    \label{fig:sup}
\end{figure*}

\section{RING-NeRF}
\subsection{Architecture details}

While our architecture presents the advantage of being quite simple and able to be adapted to different tasks, there are a few subtle technical choices that shouldn't be overlooked, especially the decoder's details.

As illustrated in the figure \ref{fig:overview} of the article, the final feature after multi-resolution combination is passed into a normalization layer to improve convergence stability. Novel View Synthesis experiments showed slightly better results when using standard, non-learnable normalization over the feature. Because a summation replaces the concatenation usually seen in grid-based architectures, the feature is shorter and could face problems for expressing high frequencies. It is thus projected into a higher dimension space with a Random Fourier Feature mapping\footnote{M. Tancik, P. Srinivasan, B. Mildenhall, S. Fridovich-Keil, N. Raghavan, U. Singhal, R. Ramamoorthi, Ravi, J. Barron and R. Ng, Fourier features let networks learn high frequency functions in low dimensional domains, in: Advances in Neural Information Processing Systems, 2020.}. This consists in a learnable frequency filter, for which we chose a sinus filter : $y = sin(Wx)$ with $W$ a linear matrix (without bias). A linear layer transforms the filter output into both density (or SDF) and one color feature. This feature, concatenated with the direction of observation encoded into Spherical Harmonics, is fed to a MLP of 3 layers that predicts the radiance.

Following the baselines' protocol of the different evaluated tasks, we also did not use any appearance embedding for all the experiments.

\subsection{Differences with other architectures}
\label{sec:dif}

Several works intended to modify the initial architecture of grid-based NeRF. However, our architecture differs from them in several respects which have a major impact on the architecture abilities.

\begin{figure*}[!ht]
    \captionsetup{type=figure}
    \includegraphics[width=\textwidth]{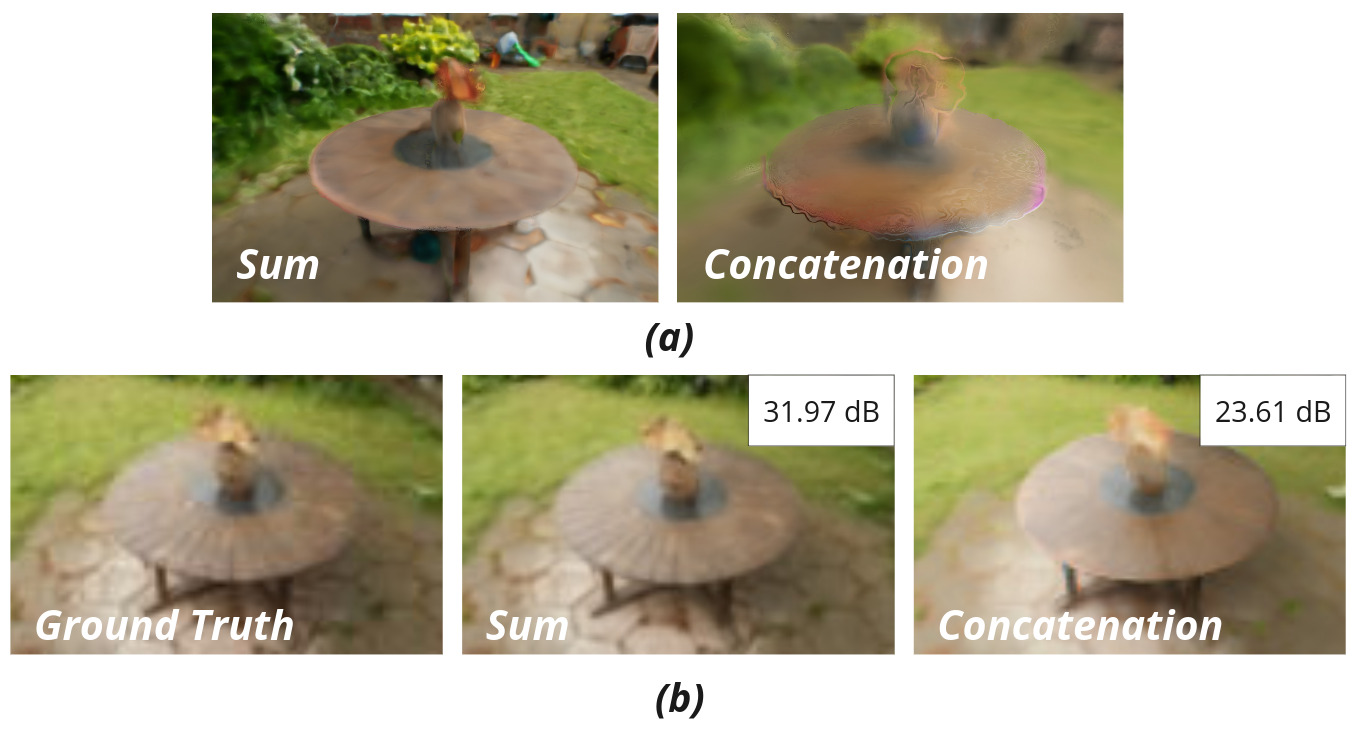}
    \captionof{figure}{Qualitative Ablative Experiment when using concatenation or sum of the features in the RING-NeRF architecture. (a) shows unsupervised LOD ($L=4$) where, although the concatenation presents semi-coherent results, they are far less qualitative than the sum's results. (b) illustrates the impact on the full-scale training and 1/8th-scale testing experiment, with closer-to-ground-truth results with the full RING-NeRF architecture.}
    \label{fig:vrnerf}
\end{figure*}

A first noticeable difference is related to the notion of LOD that is induced by the architecture itself, independently to any supervision. This property is justified through the interpretation of the grid as a mapping function from the scene space to the decoder latent space and is also validated experimentally (see section \ref{sec:archi} and  figure \ref{fig:lod}).  Except TriMip-RF \cite{hu2023tri} and LoD-NeuS \cite{zhuang2023anti}  which possess such property but for tri-plan, other architectures need an explicit supervision to capture the notion of LOD (see ResidualMFN \footnote{S. Shekarforoush, D. Lindell, D. Fleet and M. Brubaker, Residual Multiplicative Filter Networks for Multiscale Reconstruction, in: Advances in Neural Information Processing Systems, 2022.} for instance). This property is probably one of the main reason of the convergence robustness of RING-NeRF, as  observed  through few-view reconstruction and  SDF reconstruction experiments. It also simplifies the setup of a coarse-to-fine reconstruction process and a distance-aware forward mapping without scarifying speed since no convolution \cite{zhuang2023anti} nor super-sampling \cite{barron2023zip} is implied. It is also to be noted that this architecture-induced property is especially designed by RING-NeRF and other close methods such as VR-NeRF \cite{xu2023VR-NeRF} will not present intrinsic LOD as coherent as RING-NeRF, as it uses concatenation rather than sum of the multi-resolution features. Figure \ref{fig:vrnerf} shows how RING-NeRF's intrinsic LOD are more qualitative when using sum rather than concatenation of the features. Furthermore, we also demonstrate the decrease of performance the concatenation brings to the model with an example on the generalization to novel unseen observation distances.

A second  difference is the invariance of the decoder latent space towards position (unlike NGLOD \cite{takikawa2021neural}, Tri MipRF \cite{hu2023tri}, etc, that concatenate Positional Encoding with the feature extracted from the grid) but also level of detail. LOD is exclusively encoded in the grid, unlike PyNeRF \cite{turki2023pynerf}, NGLOD \cite{takikawa2021neural} or MFLOD \cite{dou2023multiplicative} that use a per LOD decoder, or Zip-NeRF \cite{barron2023zip} and  Mip-NeRF \cite{barron2021mip} which modify the feature depending on the LOD.  This property facilitates the generability of the model as illustrated in section \ref{sec:exp_extensibility}, since the learnt latent space is independent of the scale and position of the scene elements, and then favor the generalization of a (possibly) pre-trained decoder to new scenes. 


Finally, the last specificity is related to the combination of the independence of the decoder latent space with respect to the size of the grid hierarchy, and the top-down representation of the LOD. As we confirmed experimentally (see section \ref{sec:exp_extensibility} of the article), this combination makes the maximal level of detail unbounded since the size of the grid pyramid can grow up dynamically (similarly to \cite{liu2020neural}) while keeping the access to the different LODs available (unlike \cite{liu2020neural}). This property is unique and cannot be expressed by both methods that use per-LOD decoders \cite{takikawa2021neural} and methods using concatenation of the multi-resolution features such as I-NGP \cite{muller2022instant} or VR-NeRF \cite{xu2023VR-NeRF}, as they cannot increase the size of the decoder's input.

\subsection{LOD Inductive Bias}
\label{sec:exp_natural_LOD}

As discussed in section \ref{sec:method}, the inductive biases of the RING-NeRF architecture permits to naturally produce Level Of Detail of the reconstruction, even when we only use the full resolution images to supervise the full resolution LOD and not a pyramid of images supervising intermediate LOD like usual LOD NeRF architectures. This property is further illustrated on figure \ref{fig:lod} that shows more examples on several scenes of the DTU dataset. By supervising all the LOD, we observe however a correlation between the LOD used to compute the image and the level of details in the images (see figure \ref{fig:sup}). 

Other examples of unsupervised LOD in more complex scenes from the mip-360 dataset are shown in Figure \ref{fig:lod_mip}.

\subsection{Limitations of RING-NeRF}

While simple, qualitative and efficient, we can still pinpoint a few limitations of our work which could be interesting to address in future research. First of all, in the method itself, our distance-aware forward mapping may not be as physically realistic as possible. In order to better compare the volume of the projected pixel at distance $d$ with grids' cells, we chose to cast from the pixel a cubic cone and to extract a cube depending on the distance $d$. This means that, rather than sampling a true pyramid, we consider a leveled-pyramid and do not consider the growing size of the pixel inside the sampled cube. Hence, a more realistic distance-aware LOD computation would compute this more complex shape's volume rather than consider this simpler cube.

Moreover, while we carefully chose our experiments to demonstrate as much as possible the capacity and potential of our method, we may not have gone as deep as possible to showcase the implications of our work. RING-NeRF is capable of performing dynamic resolution adaptive reconstruction. We demonstrated in section \ref{sec:exp_extensibility} that adding grids a posteriori improved the final reconstruction (while not damaging the previous ones). However, this property alone has very specific use case. In order to be useful for the most of situations, the architecture should be coupled with a carefully designed stopping criterion to determine the optimal resolution of the model. This rule should explicitly decide when does the quality/efficiency ratio reaches its maximum and then stop to increase the configuration. Moreover, for maximum performance, the stopping criterion should be local rather than global, in order to define a sparse grid hierarchy which adapts itself locally to the content of the scene. However, these extensions are considered out of the scope of this article and would need further research.

\subsection{Cone Casting with Scene Contraction}
\label{sec:contract}
\textbf{Scene Contraction Function.} We use the scene contraction function from Nerfstudio's implementation \cite{tancik2023nerfstudio}, as defined originally in Mip-NeRF \cite{barron2021mip} :  
\begin{equation}
\label{eq:contract}
    contract(p) = \begin{cases}
      p & \text{if $||p|| \leq 1$}\\
      (2-\frac{1}{||p||}).\frac{p}{||p||} & \text{if $||p|| > 1$}\\
    \end{cases}       
\end{equation}

Following Nerfstudio implementation, we use $L_\infty$ norm rather than $L_2$, as it bounds the position to a cube rather than a sphere, which is convenient with grid-based representations.
Note that this contraction function bounds the scene into a cube of size 2, which is then reduced to a cube of size 1 for coding implementation reasons. This adds a factor 2 in the LOD computation formula, which is overlooked in the rest of the article for clarity reasons.
\\ \textbf{Computation of LOD L in Contracted Space.}
We defined in section \ref{sec:dist_aware_map} of the main article a formula to compute an appropriate LOD $L \in \mathbb{R}^+$ based on the grid configuration, the sample position and the image resolution, such that :

\begin{equation}
(d.c)^{3}.det(J(p)) = (\frac{1}{f^L b})^3 \iff  L = - \frac{\log{(d.c.b.\sqrt[3]{det(J(p))})}}{\log{(f)}}  
\label{eq:da_first}
\end{equation}

Based on the contraction function defined in \ref{eq:contract}, this becomes :

\begin{equation}
L = \begin{cases}
- \frac{\log{(d.c.b)}}{\log{(f)}}  & \text{if $||p|| \leq 1$}\\
    - \frac{\log{(d.c.b)}+2\log{(2-\frac{1}{||p||})}-4\log{(||p||)}}{\log{(f)}}                        & \text{if $||p|| > 1$}\\
    \end{cases}
\label{eq:final_da}
\end{equation}
\\ \textbf{Impact of Contraction and Discussion.} With this formula, we can first notice that, in the non-contracted space (inside the cube of size 1), the LOD $L$ does not depend on the contraction function, and its value decreases with the distance as expected. 

\begin{figure*}
    \captionsetup{type=figure}
    \includegraphics[width=\textwidth]{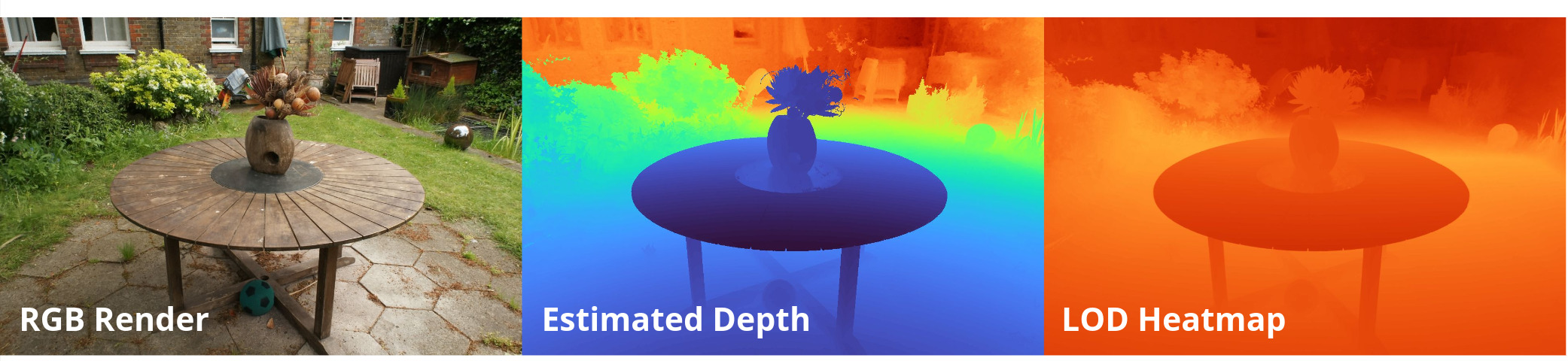}
    \captionof{figure}{Illustration of the effect of the contraction function on the chosen LOD. The depth is normalized and, in the LOD heatmap, the more intense the color, the higher the chosen LOD is.}
    \label{fig:contract}
\end{figure*}

However, further in the scene, inside the contracted space, we notice that the LOD L might increase with the distance (due to $log(||p||^4)$), as illustrated in figure \ref{fig:contract}. While this can seem pretty non-intuitive at first, this has a logical explanation. Because of the contraction, the fixed volume of a grid cell will represent an increasing volume of the scene as the distance to the scene's center increases, up to an infinite volume. This clashes with the idea that the further our sample, the lower our chosen LOD needs to be. This duality is represented in the equation by the $log(||p||^4)-log(d)$ part, and it shows that, at some point, the contraction function will always take the advantage on the distance term as our sample grow further from the center. Hence, whatever the position of the camera, if a ray continues far enough, the chosen LOD will increase at some point because of the contraction function. This also implies that the more resolute grids are used both for near objects and further areas, which would not have been the case if we did not consider the contraction function and which provides a more optimal use of the grid hierarchy with less unused parameters.

\section{New view synthesis - Section \ref{sec:nvs}}

\subsection{Configuration}

Our Novel View Synthesis and Anti-Aliasing experiments on the 360 dataset are done on one unique configuration for our baselines Nerfacto, ZipNeRF and PyNeRF and for our model RING-NeRF, both for the mono-scale and multi-scale setups. We mostly used the configuration provided in the article ZipNeRF with few differences that we will underline in the following explanations. Each of these three baselines were trained for 25k iterations with a batch size of $2^{16}$ rays. We use the RAdam Optimizer with an initial learning rate of $1e-2$, and a cosine-decayed scheduler over the whole training to $1e-3$. 

Regarding the model itself, we also use a hash grid pyramid of 10 grids with a growth factor of 2, with grid resolutions ranging from 16 to 8192. Contrary to the original Zip-NeRF, we use 8 features per level rather than 4, but with a similar hashmap size of $128^3=2^{21}$ (however please note that the ZipNeRF experiments in this paper were done also with a feature size of 8). We chose this parameter to maximize the quality of every baseline, as it is compatible with our GPU.  
We also use proposal samplers introduced by mip-NeRF 360, and once again follow Zip-NeRF configuration : 2 rounds of sampling via 2 different grid-based proposal samplers (hashgrid pyramid + one linear layer), and one last forward process into the true model to generate the pixel values. Our 2 samplers have respectively 512 and 2048 max resolution and both use features of size 1 as the sampling process only needs density information. For simplicity of implementation and comparison, we actually use the usual I-NGP-based concatenation model for RING-NeRF's proposal samplers. 



Following Nerfacto's architecture, we also used scene contraction (as described in section \ref{sec:contract} of the supplementary materials, as well as the distortion loss in these experiments, as implemented in NerfStudio for Nerfacto, PyNeRF and RING-NeRF, and their adapted improved versions for ZipNeRF as presented in their article. We also did not use appearance embedding for any method as we also noticed a decrease in metrics when using it, as observed in Mip-NeRF 360 and ZipNeRF. Finally, for Nerfacto, PyNeRF and RING-NeRF, we did not use any additional mechanism such as the ones introduced in ZipNeRF (no novel interlevel loss, nor scale featurization, nor weight decay loss, nor Affine Generative Latent Optimization as appearance embedding, ...).

The same configurations are used for both mono-scale and multi-scale experiments.


\subsection{Result Analysis and Ablative Experiments}
\label{sec:ablnvs}

\textbf{Mono-Scale Setups.}
In order to further analyze the separate impact of our architecture and of the associated mechanisms (distance-aware forward mapping and continuous coarse-to-fine), we run ablative mono-scale experiments. Our model is thus evaluated on 4 configurations with varying setups and the corresponding results can be found in table \ref{tab:nvs_xtra}. 
First of all, we notice a gap between the results of Nerfacto (27.09, 0.779 and 0.181 for respectively PSNR, SSIM and LPIPS) and our method without the distance-aware mechanism. This means that the architecture in itself (described in section \ref{sec:archi}) enables better scene reconstruction than Nerfacto.

Moreover, the addition of the distance-aware mechanism still increases the performances significantly. While this idea was initially developed as an anti-aliasing process, this experiment demonstrates that it also benefits Novel View Synthesis in situations with little variations of distances of observation.

Finally, while our continuous coarse-to-fine brings slightly better results than when using the usual discrete coarse-to-fine, it still presents slightly lower results than when not using coarse-to-fine. While the difference is not prohibitive (especially in SSIM and LPIPS) and mainly due to randomness in the training process, this observation demonstrates that, even though using coarse-to-fine on harder setups is necessary to avoid catastrophic failure or improve consistency, as explained respectively in section \ref{sec:sdfctf} and \ref{sec:ablfewViews} of the sup. mat., it does not improve results in already stable setups for Novel View Synthesis.

Table \ref{tab:supnvs} displays the results of our model against several other baselines, including Vanilla baselines (resp. NeRF, Mip-NeRF and Mip-NeRF 360). As expected, the main baselines PyNeRF, ZipNeRF and RING-NeRF all outperforms every Vanilla architectures, both in terms of quality and speed. Mip-NeRF 360 is the only vanilla method which performs close to our baselines, and also outperforms Nerfacto (although by being $50$ times slower!).

Tables \ref{tab:ps-MoAA1}, \ref{tab:ps-MoAA2}, \ref{tab:ps-MoAA3}, \ref{tab:ps-MoAA4}, \ref{tab:ps-MoAA5} and \ref{tab:ps-MoAA6} shows per-scene results of our mono-scale experiments, evaluated on the whole pyramid of resolution (note that these tables present without coarse-to-fine results to give the most qualitative possible renderings).
\\

\begin{figure}
    \captionsetup{type=figure}
    \includegraphics[width=\columnwidth]{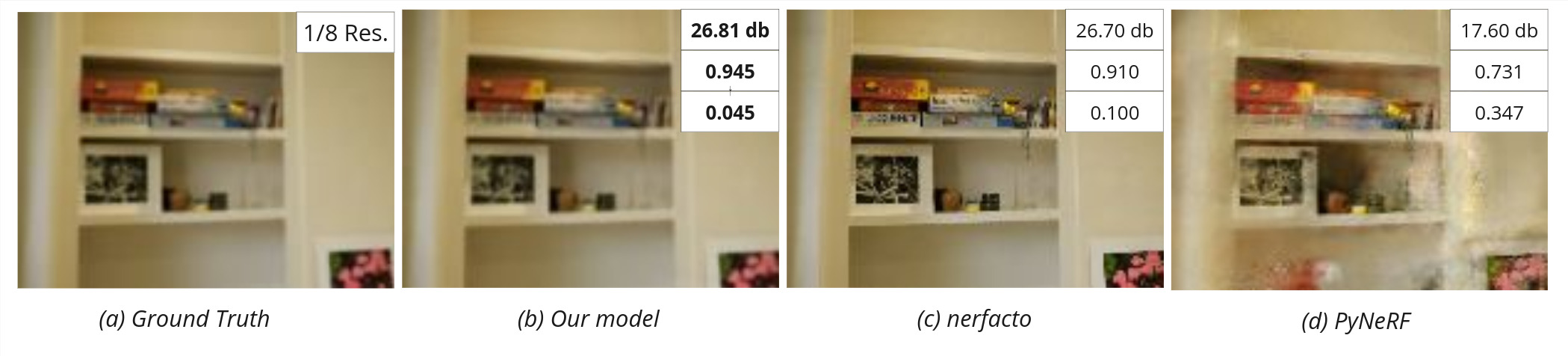}
    \captionof{figure}{Examples of instability and aliasing artifacts on novel observation distances (renders at 1/8th resolution) and impact on the metrics. Respective PSNR, SSIM and LPIPS values of the corresponding image are inset.}
    \label{fig:metrics}
\end{figure}

\begin{figure*}
    \captionsetup{type=figure}
    \includegraphics[width=\textwidth]{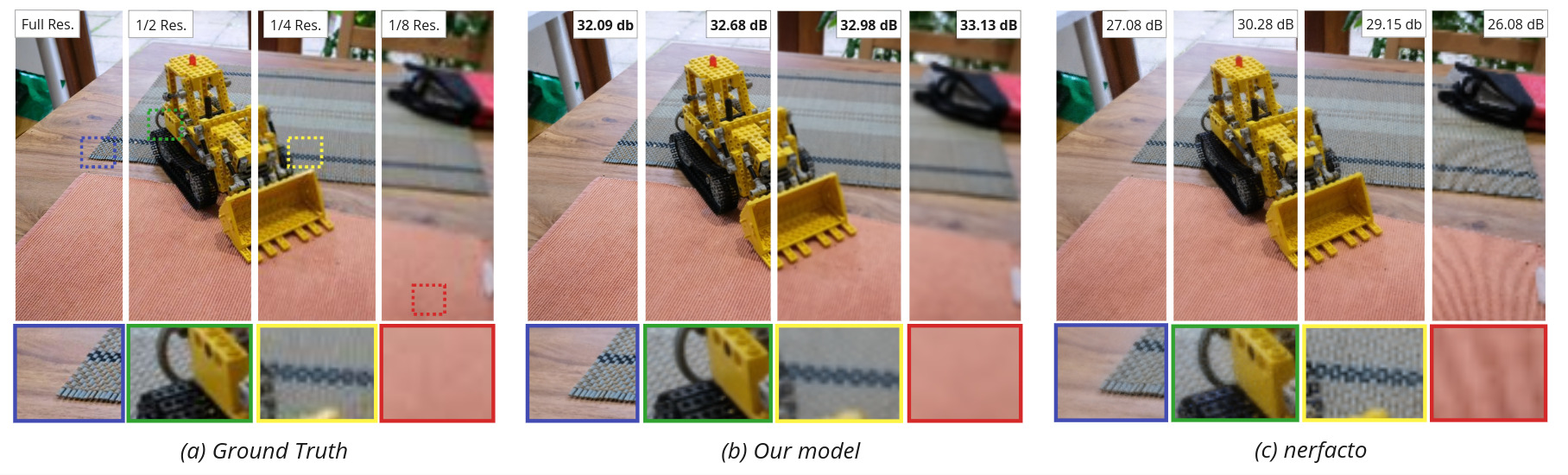}
    \captionof{figure}{Comparison of renderings of the same viewpoint with 4 different resolutions for our architecture and Nerfacto. We observe that our solution is close to the ground truth while Nerfacto faces under-contrasted image at full resolution, over-contrasted image at resolution 1/4 and aliasing artifacts at resolution 1/8. PSNR values of the corresponding image at the different resolutions are inset.}
    \label{fig:kitchen}
\end{figure*}

\textbf{Discussion on the metrics in Mono-Scale Training and Multi-Scale Testing.} When evaluating the models on novel resolutions, RING-NeRF is the only method which produces coherent rendering without aliasing artifacts, as described in section \ref{sec:nvs}. On the other hand, the other methods each behave quite differently (see table \ref{tab:nvs}). ZipNeRF produces unintelligible renderings, and thus presents the lowest quantitative results, as illustrated in figure \ref{fig:nat_aliasing}. However, while Nerfacto and PyNeRF both create similarly aliased but coherent renderings, their metrics (especially PSNR) are quite spread apart, with unexpectedly the distance-unaware Nerfacto method being better than PyNeRF. The explanation actually lies in the stability of both of these methods. As shown in figure \ref{fig:metrics}, PyNeRF often produces unstable images with novel unsupervised observation distances and viewpoints. These coarser errors have a way larger impact on metrics than the aliasing artifacts from Nerfacto, which are very localized. While this is true for every evaluation metrics, PSNR is by far the most affected by this phenomenom, as RING-NeRF is only (in average on 1/2th, 1/4th and 1/8th res. on the entire 360 dataset) $3\%$ better in PSNR than Nerfacto while they are $10\%$ and $33\%$ apart in respectively SSIM and LPIPS. On the other hand, PyNeRF and Nerfacto are $16\%$, $6\%$ and $16\%$ apart in resp. these 3 metrics (the RING/Nerfacto gap in PSNR is smaller than the Nerfacto/PyNeRF gap but higher in both SSIM and LPIPS). 
\\
\begin{table}[bt]
\caption{ Ablative study of RING-NeRF in Novel View synthesis performances for the Mono-Scale setup on the 360 dataset.}
\centering
\resizebox{0.7\columnwidth}{!}{%
\begin{tabular}{c|c c |c c c}
    & Distance-Aware  & Coarse-To-Fine             & PSNR ↑ & SSIM ↑ & LPIPS ↓  \\ \hline





Nerfacto          & - & - &  27.09    &     0.779  &     0.181 \\
RING-NeRF         &  - & - & 27.65 & 0.786    &    0.183     \\ 
RING-NeRF         & \checkmark   & - &\cellcolor{red!50} 28.26 & \cellcolor{red!50} 0.803    &   \cellcolor{red!50} 0.155        \\ 

RING-NeRF & \checkmark &  Discrete         &  \cellcolor{yellow!50} 28.06 & \cellcolor{orange!50} 0.799    &   \cellcolor{yellow!50} 0.158       \\

RING-NeRF  & \checkmark & Continuous  &  \cellcolor{orange!50} 28.09 & \cellcolor{orange!50} 0.799    &   \cellcolor{orange!50} 0.157       \\

\end{tabular}%
}

\label{tab:nvs_xtra}
\end{table}
\begin{table}[bt]
\caption{Novel View synthesis performances for the full resolution images (training and testing) on the 360 dataset.}
\centering
\resizebox{0.7\columnwidth}{!}{%
\begin{tabular}{c|c c c|c}
                 & PSNR ↑ & SSIM ↑ & LPIPS ↓ &   Training Time ↓ \\ \hline
NeRF           &  23.85    &  0.605     &  0.451     &       12.65 h        \\ 
Mip-NeRF           &   24.04   & 0.616     &   0.441    &      9.64 h          \\ 
Mip-NeRF 360       &  27.57     &     0.793   & 0.234      &        21.69 h       \\ 
\hline
Nerfacto          &   27.09    &     0.779  &     0.181   &       \cellcolor{red!50} 0.45 h     \\ 
PyNeRF           & \cellcolor{yellow!50} 27.87   &   \cellcolor{orange!50}0.802    &  \cellcolor{yellow!50}0.160     &      \cellcolor{yellow!50}    0.96 h     \\
Zip-NeRF            & \cellcolor{orange!50} 28.06   &   \cellcolor{red!50}0.808    &  \cellcolor{red!50}0.154     &      1.10 h     \\

 \hline
Our Model         &  \cellcolor{red!50} 28.09 & \cellcolor{yellow!50} 0.799    &   \cellcolor{orange!50} 0.157     &   \cellcolor{red!50}    0.45  h    \\ 
\end{tabular}%
}
\label{tab:supnvs}
\end{table}
\\
\textbf{Multi-Scale Setup.} 
Figure \ref{fig:kitchen} illustrates how anti-aliasing models behaves with multiple distances of observations against the Nerfacto baseline. On one side, the aliased rendering of Nerfacto presents different types of artifacts at different resolutions. This is due to the NeRF's model property to "average" the distances of observations between the trained views. This thus results in under-contrasted image at full resolution, over-contrasted image at resolution 1/4 and aliasing artifacts at resolution 1/8. On the other hand, RING-NeRF takes into account the distance in the computation of the feature and succeeds in adapting to the different resolutions.

Per-scene results can be found respectively in tables \ref{tab:ps-MAA1}, \ref{tab:ps-MAA2}, \ref{tab:ps-MAA3}, \ref{tab:ps-MAA4}, \ref{tab:ps-MAA5} and \ref{tab:ps-MAA6}. 

\section{Few Viewpoints Supervision - Section \ref{sec:few}}

\subsection{Configuration and Evaluation Details}
We used the same training and evaluation protocols as FreeNeRF and its predecessors PixelNeRF and Reg-NeRF. We used 15 scans among the 124 existing of the DTU \cite{jensen2014large} dataset, their IDs being : 8, 21, 30, 31, 34, 38, 40, 41, 45, 55, 63, 82, 103, 110, and 114. In each scene, the images with the following IDs are the train views: 25, 22, 28, 40, 44, 48, 0, 8, 13 (the 3, 6 and 9 views setups respectively using the first 3, 6 and 9 images). The images with IDs in [1, 2, 9, 10, 11, 12, 14, 15, 23, 24, 26, 27, 29, 30, 31, 32, 33, 34, 35, 41, 42, 43, 45, 46, 47] are the evaluation images. We also downsample each training and testing view by a factor 4, resulting in 300 x 400 pixels for each image. The masks used to evaluate the results are the ones used by FreeNeRF.

For cleaner renderings, we also decide on every of our Nerfstudio baselines (referring to the grid-based methods) to not consider any sample which is not observed by any training views when rendering novel views. This removes unnecessary noise as our model cannot imagine unseen areas.

Regarding the model, we follow FreeNeRF's choices as much as possible and, because there are no grid-based methods to compare ourselves to, the rest of the hyperparameters (mainly regarding the hashgrids) were derived from the configuration used in our Novel View Synthesis experiments. Hence, we followed the number of samples used in FreeNeRF (128 samples) and also disable the mechanisms of nerfacto dedicated to unbounded scenes : scene contraction and distortion loss. Our experiments on Nerfacto/Nerfacto+ and RING-NeRF are all done using a pyramid of 9 grids of growth factor 2, a configuration similar to our NVS experiments where we removed the last level of the highest resolution as we consider the object-centric DTU dataset simple enough to avoid to over-complexify the model. We once again use features of size 8. Based on the same observation of the dataset being simpler than the 360 dataset, we used a smaller MLP than in NVS experiments, with a hidden dimension of 64, both for density and color.  However, it is important to note that because our goal was rather to demonstrate the impact of our architecture on this task against the Nerfacto+ baseline, we did not focus our research on optimizing the configuration of the models, which could prove to further increase the metrics.

Finally, we follow FreeNeRF's density loss implementation and also use the introduced black and white prior, where we minimize the density of the $M+10$ ($5$ more samples than in FreeNeRF's implementation) first samples rather than the $M$ first when the value of the pixel is either black or white, in order to benefit from the particular form of the DTU dataset with uniform backgrounds.




\begin{table*}
\caption{Ablative Experiments on reconstruction from few viewpoints (respectively 3, 6 and 9, separated by "/") on the DTU dataset. The reported metrics are computed based on the mask of the object.}
\centering
\resizebox{\textwidth}{!}{%
\begin{tabular}{c| c | c| c c c}
                 & Density Loss  & Coarse-to-fine  & PSNR ↑ & SSIM ↑ & LPIPS ↓  \\ \hline



 

 RING-NeRF &  - &  - &   10.56 / 12.32 / 13.00  & 0.618 / 0.694 / 0.732 & 0.341 / 0.256 / 0.252  \\ 
 
  RING-NeRF & \checkmark & -  &   \cellcolor{yellow!50} 14.58 / 19.53 / 22.43 & \cellcolor{yellow!50} 0.669 / 0.780 / 0.840 & \cellcolor{yellow!50} 0.247 / 0.132 / 0.0901 \\

 RING-NeRF & -  &  Continuous  &  12.41 / 13.75 / 13.95 & 0.604 / 0.670 / 0.742 & 0.307 / 0.274 / 0.249 \\

 RING-NeRF & \checkmark & Discrete   & \cellcolor{orange!50} 15.79 / 20.16 / 22.93  & \cellcolor{orange!50} 0.706 / 0.785 / 0.847 & \cellcolor{orange!50} 0.201 / 0.127 / 0.085 \\ 
RING-NeRF  & \checkmark & Continuous    & \cellcolor{red!50} 16.18 / 20.47 / 23.19 & \cellcolor{red!50} 0.713 / 0.808 / 0.847 & \cellcolor{red!50} 0.200 / 0.127 / 0.085    \\ 
\end{tabular}%
}

\label{tab:ablfewViews}
\end{table*}
\subsection{Ablative Experiments}
\label{sec:ablfewViews}
In order to better estimate the increase of stability brought by our architecture, we further developed the ablative experiments on the complex few images setup. Table \ref{tab:ablfewViews} shows the results of different RING-NeRF versions on respectively 3, 6 and 9 images with the 15 previously stated DTU scenes. We evaluate the impact of the coarse-to-fine process (both our continuous version and the original version) and of the density loss introduced by FreeNeRF. 

\begin{figure*}
    \captionsetup{type=figure}
    \includegraphics[width=\textwidth]{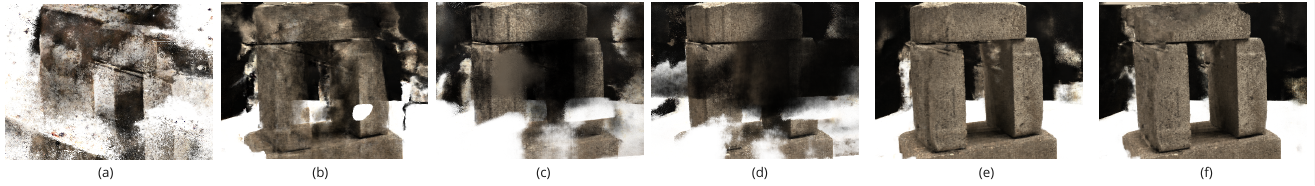}
    \captionof{figure}{Results of the Few View Ablative Experiments. (a) nerfacto (b) nerfacto w/ density loss (c) RING-NeRF w/o both density loss and continuous coarse-to-fine (d) RING-NeRF w/ continuous coarse-to-fine and w/o density loss (e) RING-NeRF w/ density loss and w/o continuous coarse-to-fine (f) RING-NeRF w/ both continuous coarse-to-fine and density loss. The model is trained on 9 images.}
    \label{fig:ablfv}
\end{figure*}

First of all, the results without both of these mechanisms are, as expected, quite low. However, we notice that its results are in average better than the basic Nerfacto architecture, especially when training with 6 and 9 views. While both are low results, we see qualitatively (see figure \ref{fig:ablfv}) that the sole RING-NeRF architecture mostly succeeds in creating a coherent 3D geometry, although not perfect and most of all very affected by floaters artifacts.  This proves that our RING-NeRF architecture is more stable than the concatenation architecture by design, even though it is still lacking of a way to get rid of artefacts-inducing ambiguities. 

\begin{figure*}
\centering
    \captionsetup{type=figure}
    \includegraphics[width=.6\columnwidth]{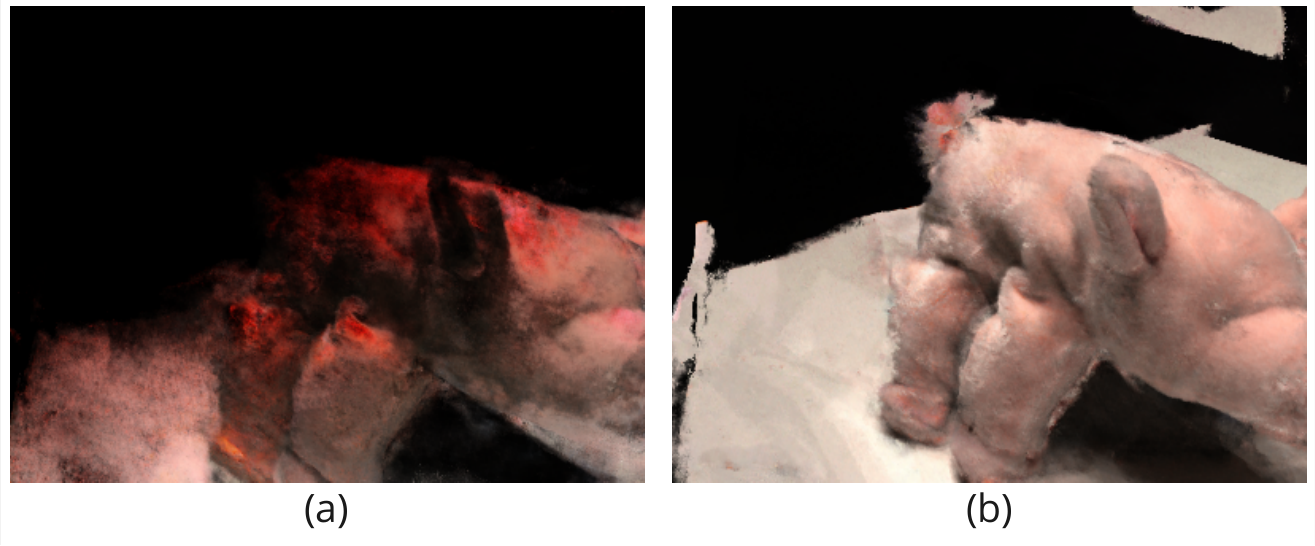}
    \captionof{figure}{Example of Inconsistency when removing the continuous coarse-to-fine from RING-NeRF. (a) RING-NeRF w/ density loss (b) RING-NeRF w/ continuous coarse-to-fine and density loss. The model is trained with 3 images that are far from the evaluated view.}
    \label{fig:ablfv2}
\end{figure*}

We also evaluate the model both without coarse-to-fine and without the density loss separately. Using solely the density loss surprisingly results in correct metrics. However, the visualization demonstrates that the 3D reconstruction in itself is slightly less precise than when using only coarse-to-fine, although the metrics are better as they are more impacted by the artefacts.  Moreover, the reconstruction without coarse-to-fine is less stable, with worse coherency in very hard setups, such as when observing the scene with only 3 views and with novel views very far from the training views, as illustrated in figure \ref{fig:ablfv2}. Note that, to further demonstrate that the density loss is not the sole explanation of the results, figure \ref{fig:ablfv} (b) shows the results of the nerfacto architecture coupled with the density loss, and we notice that, while a beginning of 3D consistency appears, it is still way more approximate than its RING-NeRF counterpart (e).

These experiments thus demonstrate that our model, coupled with continuous coarse-to-fine is able to create a coherent and qualitative 3D reconstruction, although surrounded by a huge amount of background misplaces, which have a huge impact on the metrics. The simple density loss is an efficient solution to counter this issue, but without continuous coarse-to-fine, it will in return slightly degrade the reconstruction, as well as loose in stability. The combination of both is thus the best trade off between 3D reconstruction and NVS quality.

\section{SDF Reconstruction - Section \ref{sec:sdf}}


\subsection{Choice of the baselines and Implementation}

We chose NeuS-facto and Neuralangelo as main comparing baselines. The first one is considered a fast and efficient Python SDF-based implementation and the latter is considered state-of-the-art for surface reconstruction methods. 


While NeuS-facto is solely implemented in the Sdfstudio \cite{Yu2022SDFStudio} framework (which derives from Nerfstudio), Neuralangelo's code has been officially released and also possesses a Sdfstudio implementation. It is to be noted that, while the specificities of the article are coded in a similar fashion, both implementations make many different choices on NeRF specificities. For instance, a major difference between both frameworks is the way they sample rays at each iteration. While SdfStudio randomly chooses pixels among all the images (as in Nerfstudio), Neuralangelo follows I-NGP framework by sampling every pixels of $n$ images ($n$ being the batch size). Both solutions are valid but they enable different behaviors.  In order to harmonize as much as possible our different experiments, we decided to use the Sdfstudio implementation, as the major part of the framework is common with Nerfstudio, which was used in the other experiments.



\subsection{Configuration Details}

Our experiments are all done on the Sdfstudio framework, using the provided replica subset dataset. We use all the scenes, except the scan5, for which we experienced failure of training for every method without initialization, and expect the darker environment to heavily complexify the reconstruction in an already hard initialization-lacking setup. The estimated depths are provided by Sdfstudio and  predicted via a pre-trained OmniData\footnote{A. Eftekhar, A. Sax, J. Malik, and A. Zamir, Omnidata: A scalable pipeline for making multi-task mid-level vision datasets from 3d scans. In Proc. of the IEEE International Conf. on Computer Vision (ICCV), 2021} model, they serve as an indication rather than true ground truths.

We mostly follow the basic Neuralangelo configuration implemented by Sdfstudio. This results in 16 levels of resolutions 32 to 4096 with features of size 8 and a hashmap size of $2^{19}$. The main differences regards the scheduling of the training, initially implemented to last 500k epochs. However, we noticed a similar convergence when scaling down the scheduling parameters for the training to last 100k epochs. Hence, we decided to evaluate all three methods (RING-NeRF, NeuS-facto and Neuralangelo) on 100K epochs (with accordingly scaled down schedulers and coarse-to-fine duration). Because the replica dataset is a synthetic indoor dataset, we remove background MLPs, appearance embedding and scene contraction from all evaluated baselines.

In the following additional experiments, we provide more results without initialization, as well as results with correct scene-specific initialization of the scene using SDF inverted spheres. Note that because RING-NeRF does not input the position on its MLP, it results in a different way to code the initialization. NeuS-facto and Neuralangelo both hard code the SDF inverted sphere in the weights of the MLP, benefiting from the position input while we initialize the model randomly and then overfit the network for 1000 epochs to the required initialization scheme (inverted sphere) before beginning the "true" training on the scene.



\subsection{Results}
\label{sec:sdfctf}

\begin{figure*}
\centering
    \captionsetup{type=figure}
    \includegraphics[width=0.9\columnwidth]{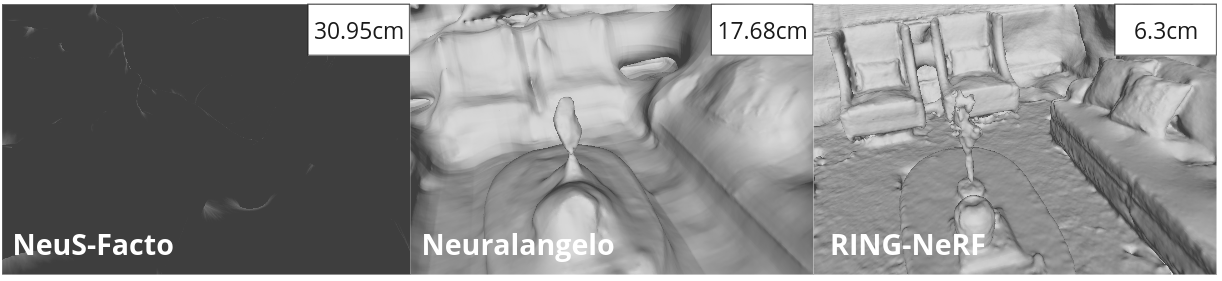}
    \captionof{figure}{Resulting Meshes for models without initialization. Chamfer Distances in cm are inset.}
    \label{fig:sdfmesh}
\end{figure*}

\begin{figure*}
    \captionsetup{type=figure}
    \includegraphics[width=\textwidth]{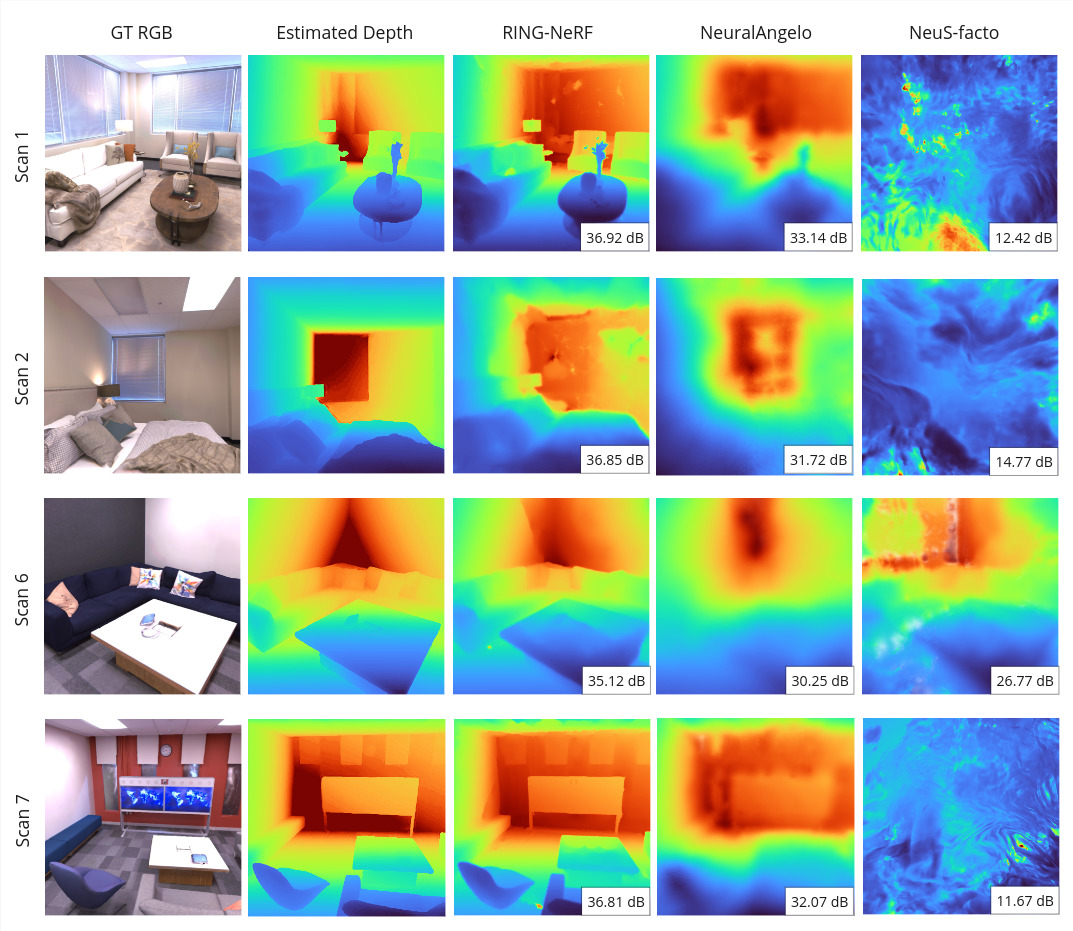}
    \captionof{figure}{Examples of Depth Prediction of different SDF-based methods while foregoing the initialization in different Replica scenes. PSNR values are the average values of the evaluation images of the entire scene.}
    \label{fig:sdf}
\end{figure*}

Additional results of SDF training without initialization on several Replica scenes can be found in figure \ref{fig:sdf}. While most training of NeuS-facto without initialization results in catastrophic failure, some scenes still succeed in producing a coherent reconstruction. The scan 6, shown in figure \ref{fig:sdf}, is one of those scenes, and while NeuS-facto avoids failure, the results are still way noisier than our RING-NeRF both in RGB and depth, and results in way lower PSNR metrics. Similarly, RING-NeRF without coarse-to-fine will often face failure, and therefore needs its continuous coarse-to-fine mechanism to forego the initialization.\\
\textbf{Reconstruction Results.} Figure \ref{fig:sdfmesh} shows some meshes obtained from the models with random initialization. As expected, the NeuS-facto mesh is completely useless, with the cloudy artefacts from the model resulting in opaque matter. On the other hand, both NeuralAngelo and RING-NeRF produces coherent meshes. However, the NeuralAngelo mesh is very blurry and contains very little details, while ours is way more precise, although rather noisy. These qualitative results are coherent with the previously described results.  \\

\begin{figure}
\centering
    \captionsetup{type=figure}
    \includegraphics[width=0.8\columnwidth]{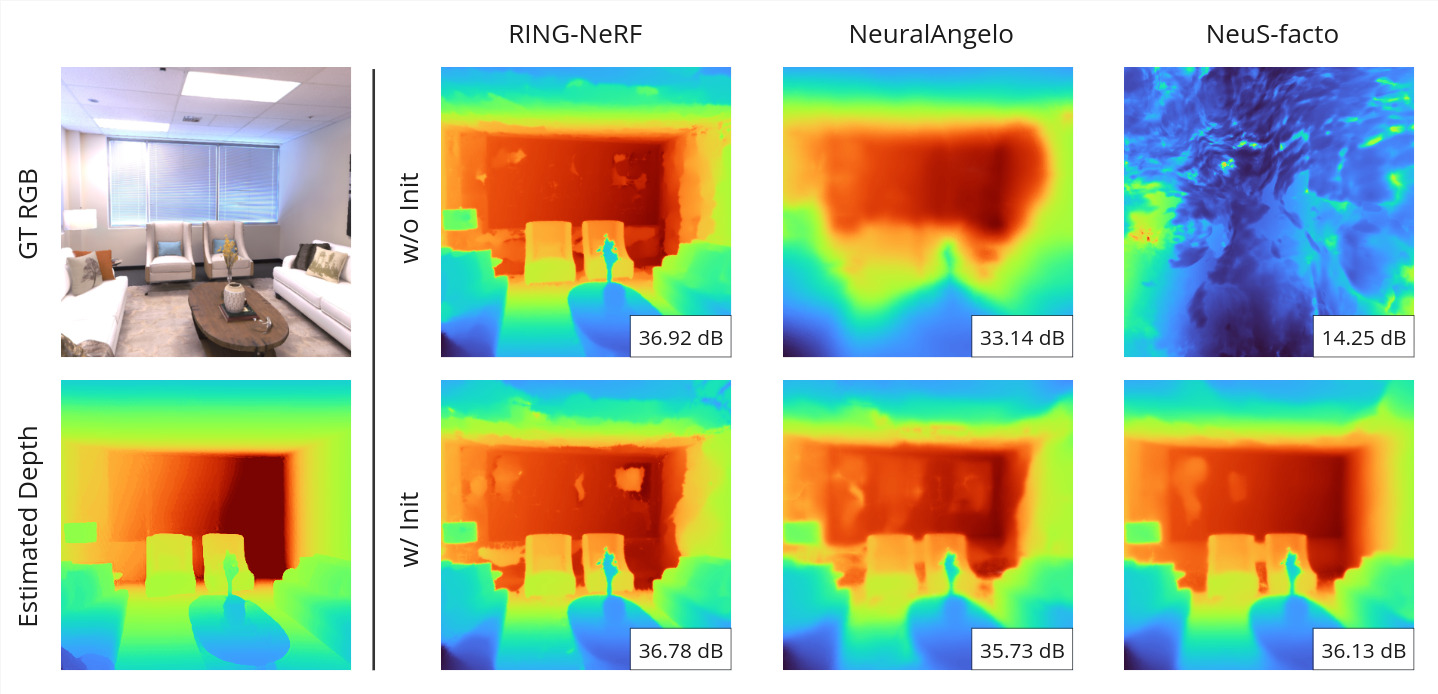}
    \captionof{figure}{Example of results with initialization.}
    \label{fig:sdfinit}
\end{figure}

\textbf{Results with initialization.}
While not the subject of the initial experiment, it is interesting to compare those results with their correctly initialized counterparts. Figure \ref{fig:sdfinit} gives an example on one scene of the dataset.

First of all, our approach presents similar results whether we initialize the model or not. This further proves that RING-NeRF is robust to the initialization process. We can notice a slight difference in the PSNR metric which can be considered within the randomness interval of two reconstruction reruns of the same model and scene.


NeuS-facto presents an expected behavior as it does not face catastrophic failure anymore and succeeds in obtaining coherent RGB renders. However, the depth remains a bit blurrier than RING-NeRF and the final PSNR results on the associated RGB images are thus affected and remains slightly lower than RING-NeRF. 


Regarding Neuralangelo, initializing the model improves both the RGB metrics and the depth estimation. However, the depth is still surprisingly blurry, which in turns prevents the RGB renders to overcome RING-NeRF results. This may be caused by the curvature loss, which is supposed to smooth the reconstruction. Because RING-NeRF introduces a bit of noise in the reconstruction, combining thoses two contributions could be an interesting research focus for robust and smooth SDF reconstruction.

\begin{figure}
\centering
    \captionsetup{type=figure}
    \includegraphics[width=0.8\columnwidth]{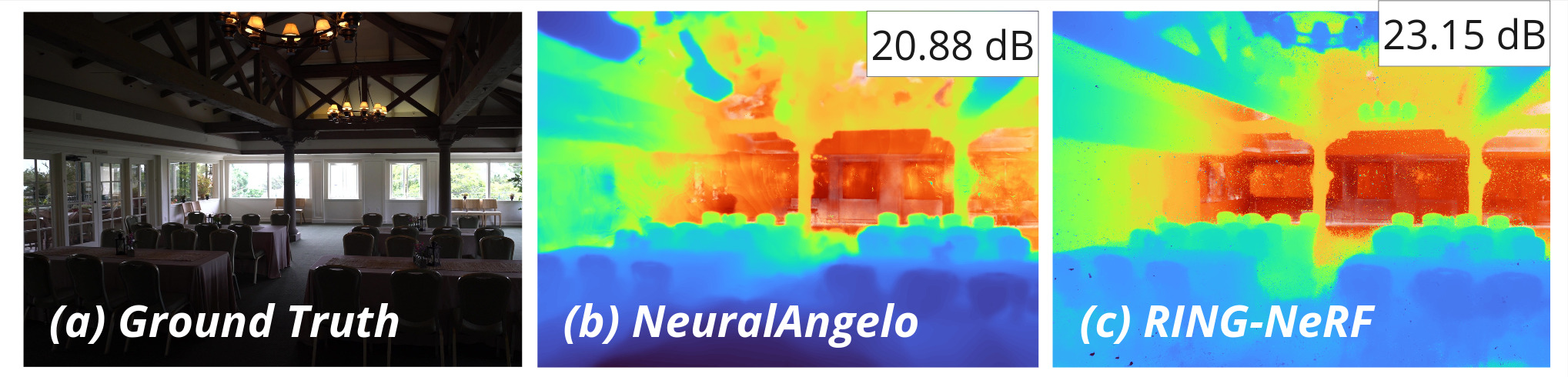}
    \captionof{figure}{Depth prediction of SDF reconstructions without initialization on one Tanks and Temple scene. Note that NeuS-Facto fails to reconstruct the scene. PSNR mean values on the test set are inset.}
    \label{fig:tt_sdf}
\end{figure}

\textbf{Results on real data.}
We demonstrated that the scene-specific initialization was crucial for other methods on the Replica dataset, contrary to our RING-NeRF. Replica being a perfect synthetic dataset, this further highlights the importance of the initialization in SDF representation. However, we also performed this no-initialization experiment on more complex real data such as the Tanks and Temples dataset. Figure \ref{fig:tt_sdf} shows results on one such scene ("Meeting Room"). As expected, while NeuS-Facto faces catastrophic failure, both NeuralAngelo and RING-NeRF succeeds in reconstructing the scene, although with varying issues. NeuralAngelo is both blurrier and with more coarse errors (eg. the ceiling's beams) while RING-NeRF faces some noise. The latter however presents better global results as confirmed by the PSNR values which are even out on a test images subset of the scene.

\section{Resolution Extensibility - Section \ref{sec:exp_extensibility}}
\subsection{Demontration of the resolution extensibility on an unbounded complex scene.}
\begin{figure}
\centering
    \captionsetup{type=figure}
    \includegraphics[width=\columnwidth]{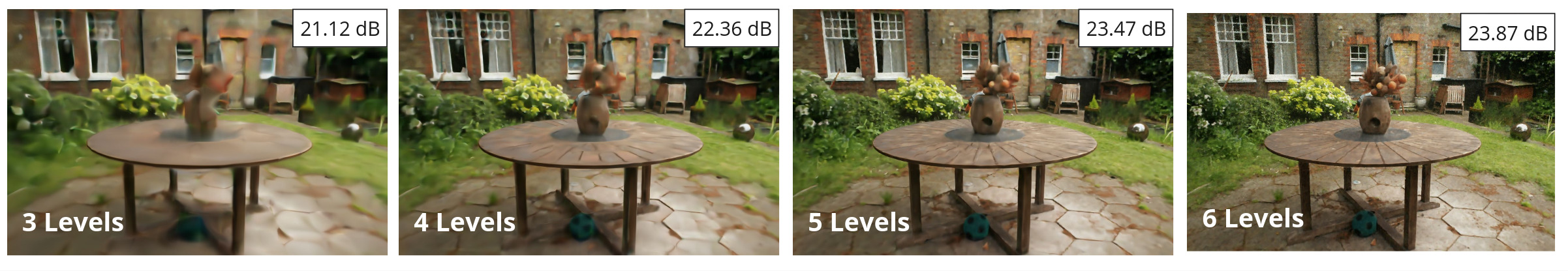}
    \captionof{figure}{Resolution Extensibility on the Garden scene of the 360 dataset. Apart from the first grid, each novel grid is trained alone, with both the previous grids and the decoder frozen.}
    \label{fig:garden_ext}
\end{figure}

Since the resolution extensibility is an intrinsic property of RING-NeRF, the experiment can be reproduced on any scene or configuration. Here, we demonstrate the property on a more complex scene of the 360 dataset, namely Garden. We follow the same experimental protocol described in section \ref{sec:exp_extensibility} with an increased hierarchy (6 levels from 16 to 512 max resolution). We begin the reconstruction with only the first three levels, train both the grid and the decoder to convergence and then freeze both of them. We then proceed to train to convergence one novel grid at a time, freezing the previous one at convergence. Figure \ref{fig:garden_ext} shows the final reconstruction at different output of the grid hierarchy. We notice that adding a grid always improves the reconstruction (with an increasing PSNR). This showcases both the capacity of our model to reconstruct finer details after the decoder's training and its ability to keep the coarser LOD valid. 

\subsection{Discussion on the Utility of the resolution extensibility.}
While resolution extensibility has not been heavily studied in previous works, it actually is a crucial property in several different use cases and applications. On one side, this can be useful for better compression of NeRF models. An important issue in grid-based NeRFs is the choice of hyper-parameters, and more especially the maximum resolution of the grid pyramid. Being able to dynamically change the resolution of your model gives the possibility to dynamically choose the optimal local maximum resolution during training depending on the scene's complexity, therefore discarding any useless parameters for a more optimal model. On the other side, the usefulness becomes obligation for embedded systems with important resources limitations. We can for instance imagine two types of limitations in SLAM situations, where the robot is moving without interruption:
\begin{itemize}
    \item  Time limitation : Because it is constantly discovering new environment to reconstruct, it does not have enough time to reconstruct with precision the previously seen areas. However, RING-NeRF will let the robot optimizes with higher resolution previous areas later on when it will have more resources available. 
    \item Memory Limitation : As the scene to reconstruct is of unknown size and the system possesses limited memory, it can not afford to reconstruct the entirety of the scene with maximum precision. A more viable strategy would be to reconstruct coarsely the majority of the scene and only reconstruct finer details on important areas (whether the most looked at places or those with the most objects for instance).
\end{itemize}






\begin{figure*}
    \captionsetup{type=figure}
    \includegraphics[width=\textwidth]{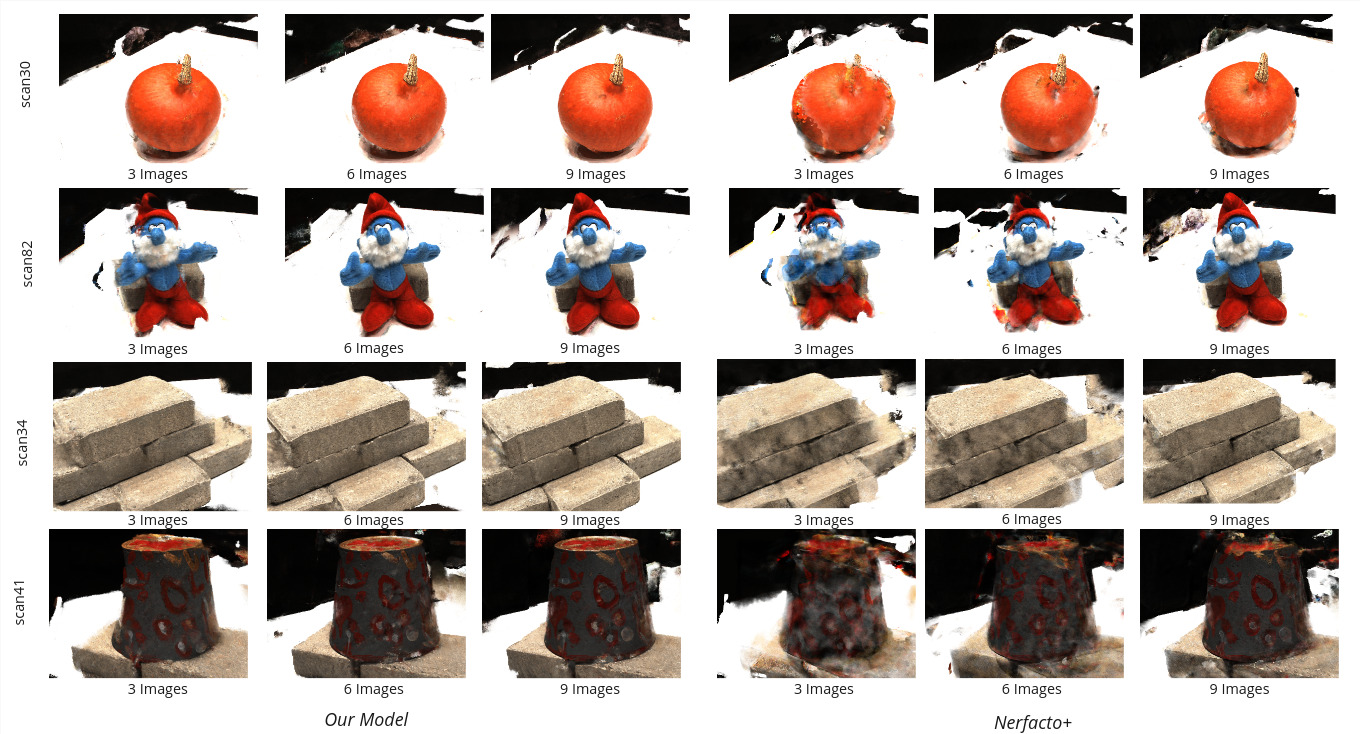}
    \captionof{figure}{Few Views experiments examples on different scans of DTU dataset.}
    \label{fig:fewres}
\end{figure*}


\clearpage

\clearpage


\begin{table*}[tb]
\caption{ Novel View Synthesis PSNR for the \textbf{Mono-Scale} setup per scene on the 360 Dataset (outside scenes).}
\centering
\resizebox{\textwidth}{!}{%
\begin{tabular}{c|cccc|cccc|cccc|cccc|cccc}
        PSNR  ↑ & \multicolumn{4}{c|}{Bicycle}  & \multicolumn{4}{c|}{Flowers} & \multicolumn{4}{c|}{Garden} & \multicolumn{4}{c|}{Stump} & \multicolumn{4}{c}{Treehill}      \\
     Res  & x1 & x2 & x4 & x8 & x1 & x2 & x4 & x8 & x1 & x2 & x4 & x8 & x1 & x2 & x4 & x8 & x1 & x2 & x4 & x8 \\ \hline
     
Nerfacto  &  24.75 & \cellcolor{orange!50}25.84 &\cellcolor{orange!50} 23.84 &\cellcolor{orange!50} 22.08 & 21.43 & \cellcolor{orange!50} 22.54 & \cellcolor{orange!50} 21.99 & \cellcolor{orange!50} 20.26 & 27.05 & \cellcolor{orange!50} 26.41 & \cellcolor{orange!50} 23.91 &\cellcolor{orange!50}  22.20 & 26.11 & \cellcolor{orange!50} 23.37 & \cellcolor{orange!50} 23.38 & \cellcolor{orange!50} 22.90 & 22.71 & \cellcolor{red!50} 24.99 & \cellcolor{orange!50} 24.58 & \cellcolor{orange!50} 23.41 \\
PyNeRF  &  \cellcolor{red!50} 25.46 & \cellcolor{yellow!50} 22.95 & \cellcolor{yellow!50} 22.37 &\cellcolor{yellow!50}  21.25 & \cellcolor{red!50}21.93 & \cellcolor{yellow!50} 17.73 & \cellcolor{yellow!50} 16.65 & \cellcolor{yellow!50} 16.55 & \cellcolor{red!50} 28.55 &\cellcolor{yellow!50} 22.96 & \cellcolor{yellow!50} 21.75 & \cellcolor{yellow!50} 20.71 & \cellcolor{yellow!50} 26.40 & \cellcolor{yellow!50} 21.65 & \cellcolor{yellow!50} 21.40 & \cellcolor{yellow!50} 20.87 & \cellcolor{orange!50} 23.28 &\cellcolor{yellow!50}  18.50 &\cellcolor{yellow!50} 17.89 &\cellcolor{yellow!50} 17.60 \\
\end{tabular}

     

}

\label{tab:ps-MoAA1}
\end{table*}

\begin{table*}[tb]
\caption{ Novel View Synthesis PSNR for the \textbf{Mono-Scale} setup per scene on the 360 Dataset (inside scenes).}
\centering
\resizebox{\textwidth}{!}{%
\begin{tabular}{c|cccc|cccc|cccc|cccc}
        PSNR  ↑ & \multicolumn{4}{c|}{Room}  & \multicolumn{4}{c|}{Counter} & \multicolumn{4}{c|}{Kitchen} & \multicolumn{4}{c}{Bonsai}                                                                                                                                                                                                                                              \\
     Res  & x1 & x2 & x4 & x8 & x1 & x2 & x4 & x8 & x1 & x2 & x4 & x8 & x1 & x2 & x4 & x8  \\ \hline

Nerfacto  &  31.59 & \cellcolor{red!50} 29.77 & \cellcolor{red!50} 29.44 & \cellcolor{red!50} 26.58 & 26.33 & \cellcolor{orange!50} 27.18 & \cellcolor{red!50}25.79 &\cellcolor{orange!50} 23.81 & 30.54 & \cellcolor{red!50} 29.70 & \cellcolor{orange!50}25.95 & \cellcolor{orange!50}23.46 &\cellcolor{yellow!50} 33.37 & \cellcolor{orange!50} 31.45 & \cellcolor{orange!50} 28.23 & \cellcolor{orange!50} 25.24\\

PyNeRF  &  \cellcolor{yellow!50} 32.12 & \cellcolor{yellow!50}24.51 &\cellcolor{yellow!50} 23.52 & \cellcolor{yellow!50} 22.63 & \cellcolor{yellow!50} 27.76 & \cellcolor{yellow!50} 19.38 & \cellcolor{yellow!50} 18.76 &\cellcolor{yellow!50} 18.52 & \cellcolor{orange!50} 32.56 & \cellcolor{yellow!50} 25.35 &\cellcolor{yellow!50}24.25 & \cellcolor{yellow!50} 22.86 & 32.80 & \cellcolor{yellow!50}24.41 & \cellcolor{yellow!50}23.73 & \cellcolor{yellow!50}22.48 \\
\end{tabular}


}

\label{tab:ps-MoAA2}
\end{table*}

\begin{table*}[tb]
\caption{ Novel View Synthesis SSIM for the \textbf{Mono-Scale} setup per scene on the 360 Dataset (outside scenes).}
\centering
\resizebox{\textwidth}{!}{%
\begin{tabular}{c|cccc|cccc|cccc|cccc|cccc}
        SSIM  ↑ & \multicolumn{4}{c|}{Bicycle}  & \multicolumn{4}{c|}{Flowers} & \multicolumn{4}{c|}{Garden} & \multicolumn{4}{c|}{Stump} & \multicolumn{4}{c}{Treehill}      \\
     Res  & x1 & x2 & x4 & x8 & x1 & x2 & x4 & x8 & x1 & x2 & x4 & x8 & x1 & x2 & x4 & x8 & x1 & x2 & x4 & x8 \\ \hline

Nerfacto  & 0.694 & \cellcolor{red!50} 0.774 & \cellcolor{yellow!50}0.662 &\cellcolor{yellow!50} 0.560 & 0.579 & \cellcolor{orange!50}0.661 & \cellcolor{orange!50}0.645 & \cellcolor{orange!50} 0.551 & 0.819 & \cellcolor{orange!50}0.767 & \cellcolor{yellow!50} 0.632 & \cellcolor{yellow!50} 0.554 & 0.730 & 0.613 & \cellcolor{yellow!50}0.555 &\cellcolor{yellow!50} 0.481 & 0.590 & \cellcolor{red!50}0.698 &\cellcolor{orange!50} 0.683 & \cellcolor{orange!50}0.647  \\

PyNeRF  &  \cellcolor{orange!50}0.726 & \cellcolor{yellow!50} 0.734 & \cellcolor{orange!50}0.682 & \cellcolor{orange!50}0.613 & \cellcolor{yellow!50} 0.592 & 0.535 & \cellcolor{yellow!50}0.538 & \cellcolor{yellow!50}0.543 & \cellcolor{orange!50} 0.847 &\cellcolor{yellow!50} 0.748 &\cellcolor{orange!50} 0.638 & \cellcolor{orange!50}0.582 & \cellcolor{yellow!50} 0.752 &\cellcolor{yellow!50} 0.637 & \cellcolor{orange!50}0.598 & \cellcolor{orange!50}0.533 & \cellcolor{orange!50}0.632 & 0.602 & \cellcolor{yellow!50}0.599 & \cellcolor{yellow!50}0.586  \\
\end{tabular}


}

\label{tab:ps-MoAA3}
\end{table*}

\begin{table*}[tb]
\caption{ Novel View Synthesis SSIM for the \textbf{Mono-Scale} setup per scene on the 360 Dataset (inside scenes).}
\centering
\resizebox{\textwidth}{!}{%
\begin{tabular}{c|cccc|cccc|cccc|cccc}
        SSIM  ↑ & \multicolumn{4}{c|}{Room}  & \multicolumn{4}{c|}{Counter} & \multicolumn{4}{c|}{Kitchen} & \multicolumn{4}{c}{Bonsai}                                                                                                                                                                                                                                              \\
     Res  & x1 & x2 & x4 & x8 & x1 & x2 & x4 & x8 & x1 & x2 & x4 & x8 & x1 & x2 & x4 & x8  \\ \hline

Nerfacto  &  0.911 & \cellcolor{orange!50} 0.874 & \cellcolor{red!50}0.876 & \cellcolor{orange!50} 0.827 & 0.844 &\cellcolor{red!50} 0.861 & \cellcolor{orange!50} 0.808 & \cellcolor{orange!50} 0.752 & 0.907 & \cellcolor{orange!50} 0.869 & \cellcolor{yellow!50} 0.706 &\cellcolor{yellow!50} 0.606 & 0.940 & \cellcolor{orange!50} 0.922 & \cellcolor{orange!50} 0.841 & \cellcolor{orange!50} 0.749 \\

PyNeRF  &  \cellcolor{yellow!50}0.919 & \cellcolor{yellow!50}0.827 & \cellcolor{yellow!50}0.780 & \cellcolor{yellow!50}0.747 & \cellcolor{yellow!50}0.873 & 0.649 & \cellcolor{yellow!50}0.585 & \cellcolor{yellow!50}0.567 & \cellcolor{red!50} 0.930 & \cellcolor{yellow!50} 0.837 & \cellcolor{orange!50} 0.745 & \cellcolor{orange!50} 0.677 & \cellcolor{yellow!50}0.946 &\cellcolor{yellow!50} 0.852 &\cellcolor{yellow!50} 0.796 & \cellcolor{yellow!50} 0.727 \\
\end{tabular}
     

}

\label{tab:ps-MoAA4}
\end{table*}

\begin{table*}[tb]
\caption{ Novel View Synthesis LPIPS for the \textbf{Mono-Scale} setup per scene on the 360 Dataset (outside scenes).}
\centering
\resizebox{\textwidth}{!}{%
\begin{tabular}{c|cccc|cccc|cccc|cccc|cccc}
        LPIPS ↓ & \multicolumn{4}{c|}{Bicycle}  & \multicolumn{4}{c|}{Flowers} & \multicolumn{4}{c|}{Garden} & \multicolumn{4}{c|}{Stump} & \multicolumn{4}{c}{Treehill}      \\
     Res  & x1 & x2 & x4 & x8 & x1 & x2 & x4 & x8 & x1 & x2 & x4 & x8 & x1 & x2 & x4 & x8 & x1 & x2 & x4 & x8 \\ \hline

Nerfacto  &  0.212 &\cellcolor{orange!50} 0.167 & \cellcolor{yellow!50}0.281 &\cellcolor{yellow!50} 0.331 & \cellcolor{yellow!50}0.259 & \cellcolor{orange!50}0.191 & \cellcolor{orange!50}0.259 & \cellcolor{orange!50}0.332 & 0.121 & \cellcolor{orange!50}0.183 & \cellcolor{yellow!50}0.291 & \cellcolor{orange!50}0.338 & 0.192 & 0.270 & 0.370 & \cellcolor{orange!50}0.471 & 0.372 & \cellcolor{orange!50}0.265 &\cellcolor{orange!50} 0.283 & \cellcolor{orange!50}0.306 \\

PyNeRF  & \cellcolor{yellow!50}0.196 & \cellcolor{yellow!50}0.176 & \cellcolor{orange!50}0.236 & \cellcolor{orange!50} 0.287 & 0.282 & 0.278 & \cellcolor{yellow!50}0.341 & \cellcolor{yellow!50}0.341 & \cellcolor{orange!50} 0.096 &\cellcolor{orange!50} 0.183 &\cellcolor{orange!50} 0.288 & \cellcolor{yellow!50}0.349 & \cellcolor{orange!50}0.160 & \cellcolor{yellow!50}0.243 & \cellcolor{orange!50}0.334 & \cellcolor{yellow!50} 0.480 & \cellcolor{orange!50}0.305 & 0.299 & \cellcolor{yellow!50}0.310 & \cellcolor{yellow!50} 0.320 \\
\end{tabular}
}

\label{tab:ps-MoAA5}
\end{table*}

\begin{table*}[tb]
\caption{ Novel View Synthesis LPIPS for the \textbf{Mono-Scale} setup per scene on the 360 Dataset (inside scenes).}
\centering
\resizebox{\textwidth}{!}{%
\begin{tabular}{c|cccc|cccc|cccc|cccc}
        LPIPS ↓ & \multicolumn{4}{c|}{Room}  & \multicolumn{4}{c|}{Counter} & \multicolumn{4}{c|}{Kitchen} & \multicolumn{4}{c}{Bonsai}                                                                                                                                                                                                                                              \\
     Res  & x1 & x2 & x4 & x8 & x1 & x2 & x4 & x8 & x1 & x2 & x4 & x8 & x1 & x2 & x4 & x8  \\ \hline

Nerfacto  & 0.121 & \cellcolor{orange!50}0.091 &\cellcolor{orange!50} 0.120 & \cellcolor{orange!50} 0.153 & 0.187 & \cellcolor{orange!50}0.122 &\cellcolor{orange!50} 0.161 &\cellcolor{orange!50} 0.207 & 0.093 & \cellcolor{orange!50} 0.102 & \cellcolor{yellow!50} 0.198 & \cellcolor{yellow!50} 0.294 & 0.071 & \cellcolor{orange!50} 0.069 & \cellcolor{orange!50} 0.145 & \cellcolor{orange!50}0.239 \\

PyNeRF &\cellcolor{red!50} 0.111 & \cellcolor{yellow!50}0.194 & \cellcolor{yellow!50} 0.208 &\cellcolor{yellow!50} 0.207 & \cellcolor{orange!50}0.149 & 0.293 & 0.309 &\cellcolor{yellow!50} 0.322 & \cellcolor{red!50}0.076 & 0.123 & \cellcolor{orange!50} 0.182 & \cellcolor{orange!50} 0.262 & \cellcolor{yellow!50} 0.069 & \cellcolor{yellow!50} 0.114 & \cellcolor{yellow!50} 0.166 &\cellcolor{yellow!50} 0.245 \\

Zip-NeRF   & \cellcolor{red!50}0.111 & 0.215 & 0.373 & 0.405 & \cellcolor{red!50}0.146 &\cellcolor{yellow!50} 0.214 &\cellcolor{yellow!50} 0.307 & 0.351 & \cellcolor{yellow!50} 0.081 & 0.163 & 0.353 & 0.456 & \cellcolor{orange!50} 0.059 & 0.380 & 0.717 & 0.716 \\
\end{tabular}
}

\label{tab:ps-MoAA6}
\end{table*}


\begin{table*}[tb]
\caption{Novel View Synthesis PSNR for the \textbf{Multi-Scale} setup per scene on the 360 Dataset (outside scenes).}
\centering
\resizebox{\textwidth}{!}{%
\begin{tabular}{c|cccc|cccc|cccc|cccc|cccc}
        PSNR  ↑ & \multicolumn{4}{c|}{Bicycle}  & \multicolumn{4}{c|}{Flowers} & \multicolumn{4}{c|}{Garden} & \multicolumn{4}{c|}{Stump} & \multicolumn{4}{c}{Treehill}      \\
     Res  & x1 & x2 & x4 & x8 & x1 & x2 & x4 & x8 & x1 & x2 & x4 & x8 & x1 & x2 & x4 & x8 & x1 & x2 & x4 & x8 \\ \hline
     
Nerfacto  &  23.73 & 27.20 & 26.95 & 24.63 & 20.96 & 24.53 & 24.95 & 22.95 & 25.01 & 28.50 & 27.35 & 24.95 & 23.82 & 25.31 & 25.46 & 24.30 & 22.67 & 25.51 & 26.27 & 25.21 \\
\end{tabular}

     

}

\label{tab:ps-MAA1}
\end{table*}

\begin{table*}[tb]
\caption{Novel View Synthesis PSNR for the \textbf{Multi-Scale} setup per scene on the 360 Dataset (inside scenes).}
\centering
\resizebox{\textwidth}{!}{%
\begin{tabular}{c|cccc|cccc|cccc|cccc}
        PSNR  ↑ & \multicolumn{4}{c|}{Room}  & \multicolumn{4}{c|}{Counter} & \multicolumn{4}{c|}{Kitchen} & \multicolumn{4}{c}{Bonsai}                                                                                                                                                                                                                                              \\
     Res  & x1 & x2 & x4 & x8 & x1 & x2 & x4 & x8 & x1 & x2 & x4 & x8 & x1 & x2 & x4 & x8  \\ \hline

Nerfacto  &  28.09 & 29.14 & 28.71 & 27.92 & 22.82 & 23.51 & 23.77 & 23.51 & 27.85 & 30.42 & 29.96 & 26.72 & 29.89 & 31.68 & 31.00 & 27.49\\

PyNeRF  &  \cellcolor{orange!50}31.73 & \cellcolor{red!50} 32.75 & \cellcolor{yellow!50}32.71 & \cellcolor{yellow!50} 31.43 & \cellcolor{yellow!50} 26.89 & \cellcolor{yellow!50} 27.47 & \cellcolor{yellow!50} 27.72 & \cellcolor{yellow!50} 27.64 & \cellcolor{yellow!50} 30.68 & \cellcolor{yellow!50} 31.45 &\cellcolor{yellow!50} 31.89 & \cellcolor{yellow!50} 30.98 & \cellcolor{yellow!50} 32.74 & \cellcolor{yellow!50} 33.52 &\cellcolor{yellow!50} 33.63 & \cellcolor{yellow!50} 33.08 \\
\end{tabular}


}

\label{tab:ps-MAA2}
\end{table*}

\begin{table*}[tb]
\caption{Novel View Synthesis SSIM for the \textbf{Multi-Scale} setup per scene on the 360 Dataset (outside scenes).}
\centering
\resizebox{\textwidth}{!}{%
\begin{tabular}{c|cccc|cccc|cccc|cccc|cccc}
        SSIM  ↑ & \multicolumn{4}{c|}{Bicycle}  & \multicolumn{4}{c|}{Flowers} & \multicolumn{4}{c|}{Garden} & \multicolumn{4}{c|}{Stump} & \multicolumn{4}{c}{Treehill}      \\
     Res  & x1 & x2 & x4 & x8 & x1 & x2 & x4 & x8 & x1 & x2 & x4 & x8 & x1 & x2 & x4 & x8 & x1 & x2 & x4 & x8 \\ \hline
Nerfacto  &  0.564 & 0.787 & 0.812 & 0.719 & 0.478 & 0.706 & 0.761 & 0.679 & 0.647 & 0.837 & 0.806 & 0.714 & 0.597 & 0.696 & 0.698 & 0.601 & 0.511 & 0.717 & 0.782 & 0.757  \\
\end{tabular}


}

\label{tab:ps-MAA3}
\end{table*}

\begin{table*}[tb]
\caption{Novel View Synthesis SSIM for the \textbf{Multi-Scale} setup per scene on the 360 Dataset (inside scenes).}
\centering
\resizebox{\textwidth}{!}{%
\begin{tabular}{c|cccc|cccc|cccc|cccc}
        SSIM  ↑ & \multicolumn{4}{c|}{Room}  & \multicolumn{4}{c|}{Counter} & \multicolumn{4}{c|}{Kitchen} & \multicolumn{4}{c}{Bonsai}                                                                                                                                                                                                                                              \\
     Res  & x1 & x2 & x4 & x8 & x1 & x2 & x4 & x8 & x1 & x2 & x4 & x8 & x1 & x2 & x4 & x8  \\ \hline
     
Nerfacto  &  0.826 & 0.861 & 0.874 & 0.872 & 0.715 & 0.739 & 0.769 & 0.769 & 0.794 & 0.873 & 0.845 & 0.755 & 0.876 & 0.908 & 0.897 & 0.825 \\

PyNeRF & \cellcolor{yellow!50} 0.899 &\cellcolor{yellow!50} 0.917 & \cellcolor{yellow!50}0.919 & \cellcolor{yellow!50}0.923 & \cellcolor{yellow!50}0.845 & \cellcolor{red!50} 0.884 & \cellcolor{yellow!50}0.866 & \cellcolor{yellow!50}0.875 & \cellcolor{yellow!50}0.891 & \cellcolor{yellow!50}0.902 & \cellcolor{yellow!50}0.917 & \cellcolor{yellow!50}0.920 & \cellcolor{yellow!50}0.909 & \cellcolor{yellow!50}0.935 &\cellcolor{yellow!50} 0.928 & \cellcolor{orange!50}0.940 \\
\end{tabular}
     

}

\label{tab:ps-MAA4}
\end{table*}

\begin{table*}[tb]
\caption{Novel View Synthesis LPIPS for the \textbf{Multi-Scale} setup per scene on the 360 Dataset (outside scenes).}
\centering
\resizebox{\textwidth}{!}{%
\begin{tabular}{c|cccc|cccc|cccc|cccc|cccc}
        LPIPS ↓ & \multicolumn{4}{c|}{Bicycle}  & \multicolumn{4}{c|}{Flowers} & \multicolumn{4}{c|}{Garden} & \multicolumn{4}{c|}{Stump} & \multicolumn{4}{c}{Treehill}      \\
     Res  & x1 & x2 & x4 & x8 & x1 & x2 & x4 & x8 & x1 & x2 & x4 & x8 & x1 & x2 & x4 & x8 & x1 & x2 & x4 & x8 \\ \hline
Nerfacto  &  0.404 & 0.152 & 0.134 & 0.199 & 0.416 & 0.153 & 0.139 & 0.210 & 0.323 & 0.119 & 0.143 & 0.210 & 0.393 & 0.228 & 0.242 & 0.368 & 0.522 & 0.254 & 0.179 & 0.200 \\
\end{tabular}
}

\label{tab:ps-MAA5}
\end{table*}

\begin{table*}[tb]
\caption{Novel View Synthesis LPIPS for the \textbf{Multi-Scale} setup per scene on the 360 Dataset (inside scenes).}
\centering
\resizebox{\textwidth}{!}{%
\begin{tabular}{c|cccc|cccc|cccc|cccc}
        LPIPS ↓ & \multicolumn{4}{c|}{Room}  & \multicolumn{4}{c|}{Counter} & \multicolumn{4}{c|}{Kitchen} & \multicolumn{4}{c}{Bonsai}                                                                                                                                                                                                                                              \\
     Res  & x1 & x2 & x4 & x8 & x1 & x2 & x4 & x8 & x1 & x2 & x4 & x8 & x1 & x2 & x4 & x8  \\ \hline
     
Nerfacto  & 0.320 & 0.163 & 0.071 & 0.093 & 0.438 & 0.288 & 0.166 & 0.176 & 0.258 & 0.113 & 0.092 & 0.176 & 0.233 & 0.112 & 0.085 & 0.160 \\
\end{tabular}
}

\label{tab:ps-MAA6}
\end{table*}

\end{document}